\documentclass{article}

\usepackage[preprint,nonatbib]{neurips_2024}

\usepackage[utf8]{inputenc} 
\usepackage[T1]{fontenc}    
\usepackage{hyperref}       
\usepackage{url}            
\usepackage{booktabs}       
\usepackage{amsfonts}       
\usepackage{nicefrac}       
\usepackage{microtype}      
\usepackage{xcolor}         

\usepackage{tabularray}
\usepackage{graphicx}
\usepackage{subfigure}
\usepackage{hyperref}

\usepackage{amsmath}
\usepackage{amssymb}
\usepackage{mathtools}
\usepackage{amsthm}
\usepackage{braket}

\usepackage[capitalize,noabbrev]{cleveref}
\usepackage{algorithmic}
\usepackage{algorithm}
\usepackage{wrapfig}
\usepackage{multirow}
\usepackage{multicol}
\newcommand\algorithmicprocedure{\textbf{procedure}}
\newcommand{\algorithmicendprocedure}{\algorithmicend\ \algorithmicprocedure}
\makeatletter
\newcommand\PROCEDURE[3][default]{%
  \ALC@it
  \algorithmicprocedure\ \textsc{#2}(#3)%
  \ALC@com{#1}%
  \begin{ALC@prc}%
}
\newcommand\ENDPROCEDURE{%
  \end{ALC@prc}%
  \ifthenelse{\boolean{ALC@noend}}{}{%
    \ALC@it\algorithmicendprocedure
  }%
}
\newenvironment{ALC@prc}{\begin{ALC@g}}{\end{ALC@g}}
\makeatother
\title{Selective Knowledge Sharing for Personalized Federated Learning Under Capacity Heterogeneity}

\author{%
Zheng Wang$^{12}$ \quad Zheng Wang$^{2}$ \quad Zhaopeng Peng$^{12}$ \quad Zihui Wang$^{12}$ \quad Cheng Wang$^{12}$\thanks{Corresponding Author} \\
$^1$Fujian Key Laboratory of Sensing and Computing for Smart Cities\\ \quad $^2$School of Informatics, Xiamen University, China \\
\texttt{\{zwang, zhengwang, pengzhaopeng, wangziwei\}@stu.xmu.edu.cn}\\
\texttt{cwang@xmu.edu.cn}\\
}

\begin{document}

\maketitle

\begin{abstract}
Federated Learning (FL) stands to gain significant advantages from collaboratively training capacity-heterogeneous models, enabling the utilization of private data and computing power from low-capacity devices. However, the focus on personalizing capacity-heterogeneous models based on client-specific data has been limited, resulting in suboptimal local model utility, particularly for low-capacity clients. The heterogeneity in both data and device capacity poses two key challenges for model personalization: 1) accurately retaining necessary knowledge embedded within reduced submodels for each client, and 2) effectively sharing knowledge through aggregating size-varying parameters. To this end, we introduce Pa$^3$dFL, a novel framework designed to enhance local model performance by decoupling and selectively sharing knowledge among capacity-heterogeneous models. First, we decompose each layer of the model into general and personal parameters. Then, we maintain uniform sizes for the general parameters across clients and aggregate them through direct averaging. Subsequently, we employ a hyper-network to generate size-varying personal parameters for clients using learnable embeddings. Finally, we facilitate the implicit aggregation of personal parameters by aggregating client embeddings through a self-attention module. We conducted extensive experiments on three datasets to evaluate the effectiveness of Pa$^3$dFL. Our findings indicate that Pa$^3$dFL consistently outperforms baseline methods across various heterogeneity settings. Moreover, Pa$^3$dFL demonstrates competitive communication and computation efficiency compared to baseline approaches, highlighting its practicality and adaptability in adverse system conditions.

\end{abstract}

\section{Introduction}\label{sec1}
\begin{figure}[ht]
\begin{center}
\centerline{\includegraphics[width=1.0\columnwidth]{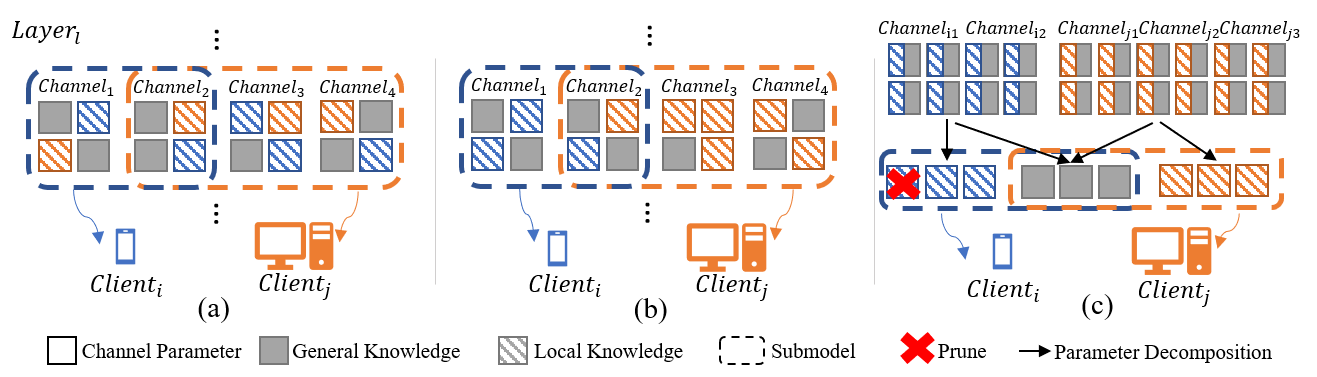}}
\caption{
$Client_i$'s smaller pruned submodel compared to $Client_j$ is due to its relatively limited capacity (e.g., mobile phone versus computer). By assuming that knowledge embedded in the model is intricately interwoven across channels within each layer, we focus on the impact of model pruning on knowledge retention at each layer for clients. \textbf{(a) Direct channel pruning} potentially loses necessary general/local knowledge from all channels and preserves irrelative knowledge from other clients. \textbf{(b) Personalized channel pruning} will lose general knowledge from unseen channels and can preserve all necessary local knowledge perhaps with irrelative knowledge from overlapping channels. \textbf{(c) Pruning after decomposition (ours)} fully maintains general knowledge through complete general parameters sharing, and local knowledge varies with pruned personal parameters}
\label{fig_example}
\end{center}
\end{figure}
Federated learning (FL) leverages private users' data on edge devices, e.g., mobile phones and smartwatches, to collaboratively train a global model without direct access to raw data \cite{mcmahan2017communication}. In practice, the clients' device capacities (i.e., hard disk or memory size) are usually heterogeneous. Some clients with limited device capacity cannot be directly arranged to train a large-size model. Therefore, traditional FL will be hindered by either limited model sizes or the absence of data from low-capacity clients \cite{mei2022resource}. To mitigate this issue, recent works \cite{li2020lotteryfl,li2021fedmask,li2021hermes,wen2022federated,diao2020heterofl,horvath2021fjord, alam2022fedrolex, deng2022tailorfl, chen2023efficient} enable collaboratively training a large-size model by reducing it into small trainable submodels. While succeeding in training capacity-heterogeneous models, the model inference performance on each client side can be hindered by both capacity and data heterogeneity. 
For one thing, the model utility will be quickly reduced as the number of model parameters decreases \cite{mei2022resource}. For another thing, the pruned submodels may not be suitable for client-specific data distributions \cite{deng2022tailorfl}. The two issues motivate us to consider such a problem:

\begin{center}
      \textit{How to train personalized FL models simultaneously based on clients' capacity and data?}
\end{center}
There have been plenty of works addressing model personalization based on client-specific data  \cite{li2021ditto, t2020personalized},
most of which demand a uniform model size being shared across clients. Personalization methods based on knowledge distillation (KD) support the training of heterogeneous models, but they fail to leverage the consistency in model parameters and architectures. A most recent step towards this problem is TailorFL \cite{deng2022tailorfl} and pFedGate\cite{chen2023efficient}, i.e. dropping out excessive channels from the global model based on clients' data and capacity. However, they may still lead to suboptimal local performance since \textbf{knowledge necessary for clients cannot be well retained when model parameters are pruned} as shown in Fig.\ref{fig_example}(a) and (b) (w.r.t.,   
general knowledge refers to what is suitable for all the clients while local knowledge is only necessary for partial clients). 
The undesired knowledge retention will harm FL effectiveness and efficiency in two ways: 1) the inconsistency between knowledge embedded in submodels and local data increases the difficulty of local model training and, 2) different channels may redundantly learn similar knowledge that cannot be aggregated within the same layer \cite{diao2020heterofl, horvath2021fjord, chen2023efficient, deng2022tailorfl, mei2022resource}. Moreover, traditional position-wise parameter averaging \cite{chen2023efficient, deng2022tailorfl} may further lower the aggregation efficiency due to unmatched fusion of knowledge independently learned by different clients (e.g., fusing global and local knowledge) \cite{wang2020matched, li2022position}. These issues induce two key challenges for capacity-heterogeneous model personalization: 
\begin{description}\label{cha}
    \item[Challenge 1.]Accurately retaining necessary knowledge when pruning the model for each client.
    \item[Challenge 2.] Effectively aggregating knowledge from capacity-heterogeneous models across clients.
\end{description}

Motivated by recent advancements in parameter decomposition techniques \cite{mei2022resource, jeong2022factorized, hyeon2021fedpara}, we introduce a novel framework, termed Pa$^3$dFL (\textbf{P}runing \textbf{a}nd \textbf{a}ggregation \textbf{a}fter \textbf{d}ecomposition \textbf{FL}), aimed at tackling the challenges outlined above. To address the first challenge, we decompose each layer's model parameters into general and personal components as depicted in Fig.\ref{fig_example}(c). Subsequently, we trim the model size by only pruning personal parameters while preserving the consistency of global knowledge embedded in general parameters across all clients. In this way, only client-specific and general knowledge will be allocated to each client within the submodel. To address the second challenge, we directly average general parameters with constant shapes to foster comprehensive general knowledge sharing. Next, we adopt a hyper-network to generate personal parameters from learnable client embeddings. Finally, we facilitate the implicit aggregation of personal parameters by aggregating client embeddings via a self-attention module, which adaptively enhances personal knowledge sharing among similar clients.

Our contributions are concluded as follows:
\begin{itemize}
\item We identify two key challenges (i.e., \textbf{Challenges} \textbf{1} and \textbf{2}) in model personalization under data and capacity heterogeneity, which are commonly overlooked by previous works.
\item We propose an efficient personalized FL framework to enhance local model performance
by decoupling and selectively sharing knowledge among capacity-heterogeneous models. 
\item We conduct extensive experiments under various heterogeneity settings to show the advantage of our proposed methods in terms of accuracy and efficiency.
\end{itemize}
\section{Problem Formulation}\label{sec3}
 In personalized FL, there exist $N$ clients equipped with their private data $\{\mathcal{D}_i| i\in [N]\}$, and the goal is to obtain a series of model $\mathcal{M}_{\Theta} = \{\mathcal{M}_{\theta_1},\mathcal{M}_{\theta_2}, ..., \mathcal{M}_{\theta_N}\}$ that minimize
 \begin{equation}
     \min_{\Theta}f(\Theta)\coloneqq \frac1N\sum_{i\in [N]}{\mathbb{E}_{(x,y)\sim \mathcal{D}_i}\left[f( \mathcal{M}_{\theta_i}; x,y)\right]} 
 \end{equation}
 where $f(\mathcal{M}_{\theta}; x,y) = \ell (\mathcal{M}_{\theta}(x), y)$ and $\ell(\cdot)$ is  the loss function. In a capacity-heterogeneous setting, the constraints on both running-time peak memory occupation and model size may be different for clients, and the local model ability will finally be limited by the weaker one of the memory term and the storage term. Following  \cite{chen2023efficient}, we denote the full-model size $s_{max} = |\mathcal{M}_{\theta_{all}}|$ and assume that each client can train a submodel of the reduced size ratio $r_i\le \frac{s_i}{s_{max}}$. Consequently, we formulate the capacity constraint as
\begin{equation}\label{eq_capacity_cons}
    |\mathcal{M}_{\theta_i}|\le r_i s_{max}, \forall i\in [N]
\end{equation}
under which the PFL objective (1) should be solved.

\section{Methodology}\label{sec4}
\subsection{Channel-aware Layer Decomposition}
We start from the decomposition of the weight of the convolution operator introduced by FLANC \cite{mei2022resource}, which can also be extended to fully connected layers by setting kernel size to $1$. Given a convolution weight tensor $\boldsymbol{\theta} \in \mathbb R^{T\times S\times k\times k}$ ( w.r.t., $T$ the output channels, $S$ the input channels, and $k$ the kernel size), the decomposition of FLANC consists of a general part $\boldsymbol{\theta_u}\in \mathbb R^{k^2R_1\times R_2}$ and a capacity-dependent part $\boldsymbol{\theta_v}\in \mathbb R^{R_2\times \frac{S}{R_1}T}$ such that
\begin{equation}\label{eqflanc}
    \boldsymbol{\theta}' = \boldsymbol{\theta_u}\boldsymbol{\theta_v}=
    \left[\underbrace{\boldsymbol{\theta_u}\boldsymbol{\theta}_{\bold v,1}}_{\boldsymbol{\theta}'_1}, \cdots,\underbrace{ \boldsymbol{\theta_u}\boldsymbol{\theta}_{\bold v,\frac{ST}{R_1}}}_{\boldsymbol{\theta}'_{\frac{ST}{R_1}}}\right] =  \left[\cdots,\underbrace{\boldsymbol{\theta}'_{\frac{S(i-1)}{R_1}+1:\frac{Si}{R_1} } }_{\text{Output Channel } o_i, i=1 \text{ to } T}, \cdots\right],\\\boldsymbol{\theta}=\text{Reshape}_{\text{FLANC}}(\boldsymbol{\theta}') 
\end{equation}
where $R_1, R_2$ are manually predefined coefficients of the decomposition and $\boldsymbol{\theta}$ is obtained by reshaping parameters according to channels in $\boldsymbol{\theta}'$. FLANC reduces the model size by only reducing the columns of $\boldsymbol{\theta_v}$ to $SpTp/R_1$ (i.e., $0< p\le 1$ is the reduction ratio of the model width) while keeping $\boldsymbol{\theta_u}$ consistent across clients. This
reveals an outstanding property that general knowledge learned by $\boldsymbol{\theta_u}$ can be propagated to all the reduced models to access the full range of knowledge on all capacity-heterogeneous devices. Motivated by FLANC, we employ a similar decomposition scheme where capacity-dependent parameters are personally kept instead of shared among devices with the same capacity. However, we notice that general parameters $\boldsymbol{\theta_u}$ in FLANC can hardly encode channel-specific knowledge since $\boldsymbol{\theta_u}$ equally contributes to every output channel as the basis in Eq. (\ref{eqflanc}). This may limit model performance, especially when output channels sensitive to certain signals \cite{zeiler2014visualizing} (e.g., low-level circles or high-level eyes) are indispensable across all clients' local tasks. 
To mitigate this issue, we propose to decompose $\boldsymbol{\theta}$ in a channel-aware manner
\begin{equation}\label{eqpad}
    \boldsymbol{\theta}' = 
    \underbrace{\left[
    \begin{matrix}
        \boldsymbol{u}_{1}\\
        \boldsymbol{u}_{2}\\
        \vdots\\
        \boldsymbol{u}_{R_1}
    \end{matrix}
    \right]}_{\boldsymbol{\theta_u}\in \mathbb R^{k^2R_1\times R_2}}
    \underbrace{\left[
    \boldsymbol{v}_{1}, \cdots, \boldsymbol{v}_{\frac{T}{R_1}}, 
    \right]}_{\boldsymbol{\theta_v}\in \mathbb R^{R_2\times \frac{T}{R_1}S}} = \underbrace{\left[
    \begin{matrix}
        \boldsymbol{u}_1\boldsymbol{v}_{1}& \cdots& \boldsymbol{u}_1\boldsymbol{v}_{\frac{T}{R_1}}\\
        \vdots&\ddots&\vdots\\
        \boldsymbol{u}_{R_1}\boldsymbol{v}_{ 1}&\cdots&\boldsymbol{u}_{R_1}\boldsymbol{v}_{\frac{T}{R_1}}
    \end{matrix}
    \right]}_{\text{Output Channel }o_{ij} = \boldsymbol{u}_{i}\boldsymbol{v}_{j}|_{i=1 \text{ to }R_1, j=1 \text{ to }\frac{T}{R_1}}},\boldsymbol{\theta}=\text{Reshape}_{\text{PadFL}}(\boldsymbol{\theta}') 
\end{equation}
where $\boldsymbol{u}_{i}\in \mathbb R^{k^2\times R_2}$, $\boldsymbol{v}_{j}\in \mathbb R^{R_2\times S}$ and 
$\boldsymbol{\theta}'$ is also reshaped by channels to obtain final $\boldsymbol{\theta}$. Eq. (\ref{eqpad}) enables $\boldsymbol{u}_{i}$ in general parameters to remember different channel-specific knowledge since it's only shared by partial channels. We adapt the model to clients' small capacity by reducing columns $\boldsymbol{v}_{j}$ in $\boldsymbol{\theta_v}$
, where only output channels $(\boldsymbol{u}_{\bold \cdot}\boldsymbol{v}_{j})$ will be pruned together. This completely preserves general parameters $\boldsymbol{\theta_u}$ in the reduced model as FLANC.
Additionally, we choose the coefficients $R_1, R_2$ adaptively by
\begin{equation}\label{eq_coef}
    R_1=Tp_{min}, R_2=\left \{
\begin{array}{ll}
   \max(\min(S,T),k^2),& \text{if layer is convolution}\\
    R_1,& \text{if layer is linear}\\
\end{array}
\right.
\end{equation}
to strike a balance between computation efficiency and performance as depicted in Sec.5.

\paragraph{Model Reduction.} Given a full-size model $\mathcal{M}_{\bold \theta}$ with $d$ layers, we represent the model parameters $\bold\theta$ by the set of decomposed parameters of all layers $\{(\bold \theta_{\bold u}^{(l)}, \bold \theta_{\bold v}^{(l)})\}|_{l=1\text{ to }d}$. Then, we layer-wisely remove columns in $\bold\theta_{\bold v}^{(l)}|_{l=1\text{ to } d}$ in descending order of column indices (i.e., Eq. (\ref{eq_prune})) to obtain the $p_i$-width model for each client $c_i$ based on capacity constraint Eq. (\ref{eq_capacity_cons}). We finally denote general pameters by $\bold\theta_{i,\bold u}=\{\bold \theta_{\bold u}^{(l)}\}|_{l=1\text{ to } d}$  and varying-size personal parameters by $\bold\theta_{i,\bold v}=\{\bold \theta_{\bold v}^{(l)}\}|_{l=1\text{ to } d}$ for each client $c_i$.
\subsection{Aggregation Mechanism}
The proposed model decomposition method enables clients to collaboratively maintain general parameters $\bold \theta_{\bold u}$ of a constant shape. As a result, $\bold \theta_{\bold u}$ can be updated by direct averaging at each aggregation round $t$
\begin{equation}
    \bold \theta_{\bold u}^{(t+1)}\leftarrow\frac{1}{|\mathcal{S}_t|}\sum_{i\in |\mathcal{S}_t|}\bold\theta_{i,\bold u}^{(t)}
\end{equation}
where $\mathcal S_t$ denotes the currently selected subset of clients. However, direct position-wise averaging on personal parameters $\{\bold\theta_{i,\bold v}\}|_{l=1\text{ to } N}$ of varying shapes may reduce the aggregation efficiency \cite{alam2022fedrolex} and cause unmatched knowledge fusion \cite{wang2020matched}. Moreover, since personal parameters are viewed as learners for personal knowledge
ge,
it's essential to propagate $\bold\theta_{i,\bold v}$ within clients that have similar local data distributions \cite{zhang2023fedala}. To this end, we employ a hyper-network (HN) to generate personal parameters from learnable client embeddings that will be aggregated to facilitate the implicit aggregation of personal parameters.
\paragraph{Parameter Generation via HN.} Personalizing model parameters through HN is first introduced by pFedHN \cite{shamsian2021personalized}. In pFedHN, the server generates and distributes the full model parameters $\boldsymbol{\theta}_i^{(t)}=h(\boldsymbol{\theta}^{(t)}_h, \bold e_i^{(t)})$ to the selected client $c_i, i\in \mathcal S_t$, w.r.t. learnable embeddings $\bold e_i$, parameterized HN $h(\boldsymbol{\theta}_h, \cdot)$ at the $t$th communication round. 
Then, the server collects the locally trained parameters $\boldsymbol{\theta}_i^{(t+1)}$ to update the hyper-network $h$ with the learning rate $\gamma$
\begin{align}
        \mathcal L_{HN} = \frac{1}{|\mathcal{S}_t|}\sum_{i\in\mathcal S_t}&\frac{1}{2}\|\boldsymbol{\theta}_i^{(t+1)}-\boldsymbol{\theta}_i^{(t)}\|_2^2 \label{eq_losshn}\\
    \bold e_i^{(t+1)} \leftarrow \bold e_i^{(t)} - \gamma\frac{\partial \mathcal L_{HN}}{\partial \bold e_i^{(t)}},&\quad
   \boldsymbol{\theta}^{(t+1)}_h \leftarrow \boldsymbol{\theta}^{(t)}_h - \gamma\frac{\partial \mathcal L_{HN}}{\partial \boldsymbol{\theta}^{(t)}_h} \label{eq_updatehn}
\end{align}
Unlike pFedHN, we use HN only to generate personal parameters $\boldsymbol{\theta}_{i, \bold v}|_{i=1}^N$  for clients. First, we input all client embeddings $\bold E=\left[\bold e_1, \cdots, \bold e_N\right]$ into the encoder to obtain $\bold E'=Encoder_{HN}(\bold E)=\left[\bold e_1', \cdots, \bold e_N'\right]$. Next, we compute the similarity between client $c_i$ and other clients by $\bold s_i=\bold E'^\top \bold e_i'$. Subsequently, we respectively generate personal parameters of each layer $ \tilde{\bold\theta}_{i,\bold v}^{(l)}$ by Eq. (\ref{eq_gen}) and prune them by Eq.\ref{eq_prune}
\begin{align}
    \underbrace{\bold e_i^{(l)} = \bold E' \text{softmax}(\frac{\bold s_i}{\tau_l})}_{\text{Aggregation}},&  \tilde{\bold\theta}_{i,\bold v}^{(l)} = Decoder_{HN,l}(\bold e_i^{(l)})=\left[\underbrace{\bold v_{11}^{(il)},\cdots,\bold v_{1S_l}^{(il)}}_{\bold v_1^{(il)}}, \cdots, \underbrace{\bold v_{\frac{T_l}{R_{1l}}1}^{(il)},\cdots,\bold v_{\frac{T_l}{R_{1l}}S_l}^{(il)}}_{\bold v_{\frac{T_l}{R_{1l}}}^{(il)}} \right]\label{eq_gen} \\
    &\bold\theta_{i,\bold v}^{(l)} = \text{Prune}(\tilde{\bold\theta}_{i,\bold v}^{(l)})=\left[\bold v_{1,1:p_iT_{l-1}}^{(il)}, \cdots,\bold v_{\frac{p_iT_l}{R_1l},1:p_iT_{l-1}}^{(il)} \right]\label{eq_prune}
\end{align}
where $T_{l},S_{l}$ the numbers of output, input channels of layer $l$, $\tau_l$ the learnable coefficient (i.e. temperature of softmax) of layer $l$, $p_i$ the target reduction ratio of the model width. In practice, $p_i$ can be approximately ensured by $p_i^2\le r_i$ (i.e., Eq. (\ref{eq_capacity_cons})) according to our complexity analysis in Sec. \ref{sec4.7}. In practice, we use the multi-layer perceptron as the encoder and decoder, where the 1-D output vectors will be reshaped to obtain the parameters.
\paragraph{Implict Aggregation.} Our Eq. (\ref{eq_gen}) can promote personal knowledge sharing among similar clients from three aspects. First, HN implicitly aggregates personal parameters in the embedding space instead of position-wise averaging in the original parameter space, allowing complex non-linear knowledge fusion across varying-size parameters. Second, the self-attention module enlarges aggregation weights between clients with larger similarities in the converted features, which adaptively promotes personal knowledge sharing among similar clients. Finally, we independently use learnable temperature coefficients for softmax aggregation in different layers, since the optimal personalization degree can vary across layers \cite{ma2022layerp} and tasks \cite{zhang2023fedala}. For example, $\tau_l\rightarrow \infty$ leads to no personalization on the layer $l$ and a small $\tau_l$ encourages a high degree of layer personalization.
\subsection{Implementation}
In Pa$^3$dFL implementation, we decompose all the layers but the last one (i.e., head). We follow \cite{oh2021fedbabu} to fix the global head without updating it during training and we maintain a local head by the HN respectively for each client \cite{arivazhagan2019personalhead}. We employ a regularization term slightly different from \cite{mei2022resource} during local training. We finally fuse the received model with the global head and the locally trained one with the local head on local validation data by linear search to decide each client's local model for testing. For the hyper-network, we use the multi-layer-perceptron (MLP) to build the encoder and decoders, and we view the dimensions of client embeddings and hidden layers as hyper-parameters. The details of the implementation are in Appendix\ref{apdx_b}.

\section{Analysis}\label{sec4.7}

\paragraph{Complexity.} Pa$^3$dFL reduces the global model along the width dimension. Supposing a model's width reduction ratio is $p$, the size of the model will be reduced by $\mathcal{O}(p^2)$ \cite{diao2020heterofl}. Therefore, both the computation costs and the communication costs can be significantly reduced by choosing a small reduction ratio. One concern is that Pa$^3$dFL will introduce additional costs in both computation and communication processes due to its parameter decomposition. We demonstrate that these additional costs are acceptable in our adaptive choices of $R_1$ and $R_2$ and thus won't significantly violate the benefits from the width-level reduction. We empirically compare the efficiency of Pa$^3$dFL with existing methods in Sec. \ref{sec5.4}, and we detail the complexity analysis in Appendix \ref{apdx_a}. 
\paragraph{Convergence.} We follow \cite{pillutla2022perconvergence} to assume a single local update per device and make the convergence analysis under commonly used assumptions of the smoothness of the objective (i.e., Assumption \ref{assmp:smoothness}) and bounded gradient variance (i.e., Assumption \ref{assmp:stoc-grad-var}).
\newtheorem{theorem}{Theorem}
\begin{theorem}[Convergence of Pa$^3$dFL]\label{theorem1}
    Suppose Assumptions~\ref{assmp:smoothness}, \ref{assmp:stoc-grad-var} hold and the step sizes $\eta$ and $\gamma$ in Pa$^3$dFL are chosen as  $\frac{2}{L_{\bold u}}$, $\frac{L_{\bold u}}{L_{\boldsymbol{\varphi}}}$ respectively, $\eta$ decays with the number of rounds $t$ and all model parameters initialized at the same point $\boldsymbol{\theta}^{(0)}$. Then, after $T$ rounds we have:
    \begin{equation}
         \frac{1}{T}(\sum_{t=0}^{T}\gamma\eta_{t}(1-\frac{\gamma\eta_{t}L_{\boldsymbol{\varphi}}}{2})\|\frac{\partial F}{\partial \boldsymbol{\varphi}^{(t)}}\|^{2} + \sum_{t=0}^{T}\eta_{t}(1-\frac{\eta_{t}L_{\boldsymbol{\varphi}}}{2})\|\frac{\partial F}{\partial \boldsymbol{\theta}_{\bold u}^{(t)}}\|^{2}) \le \frac{1}{T}(F(\boldsymbol{\theta}^{(0)}) - F^{*}) + O(\eta^{2})
    \end{equation}
    where $F^{*}$ is the lower bound of $F$.
\end{theorem}
\paragraph{Privacy.} Pa$^3$dFL won't produce more privacy concerns than fedavg \cite{mcmahan2017communication} since it only transfers model parameters between the server and clients like FedAvg without local data leaving clients' devices. Besides, client embeddings in Pa$^3$dFL on the server side are randomly initialized and updated only by the delta of model parameters, which can only reflect the similarities between clients without leaking their local data information.


\section{Related Work}\label{sec2}

\paragraph{Personalized FL.} 
While traditional FL finds a global solution that is suitable to all the clients \cite{mcmahan2017communication,jin2023feddyn}, personalized FL (PFL) aims to obtain customized models of high local utility for different clients \cite{smith2017federated,kairouz2021advances}. Existing works focus on improving the effectiveness of PFL by partial parameter sharing \cite{arivazhagan2019federated,liang2020think,husnoo2022fedrep,sun2021partialfed,li2021fedbn}, meta-learning\cite{t2020personalized,li2021ditto,fallah2020personalized}, latent representation alignment\cite{tan2022fedproto,oh2021fedbabu}, local history maintaining\cite{zhang2023fedala,li2021fedphp}, hyper-network\cite{shamsian2021personalized}, adaptive aggregation\cite{zhang2020personalized,huang2021personalized,xu2023personalized,luo2022adapt}, and knowledge decoupling\cite{chen2021bridging,zhang2023gpfl}. Although achieving high accuracy, these methods commonly assume a uniform model size to be shared \cite{deng2022tailorfl}\cite{diao2020heterofl}. Some model agnostic methods\cite{tan2022fedproto,li2019fedmd} can also mitigate the low-capacity issue, however, leading to suboptimal results without leveraging the consistency in model parameters and architectures.
\paragraph{Pruning FL.}

To enable collaboratively training capacity-heterogeneous models, several methods prune or mask model parameters to reduce the model size \cite{li2021hermes, li2021fedmask, horvath2021fjord}.  \cite{li2020lotteryfl,li2021fedmask,li2021hermes} mask model parameters to reduce the communicating model size, but cannot save computation costs during the model training phase.\cite{wen2022federated,diao2020heterofl,horvath2021fjord, alam2022fedrolex} drop channels out at each layer of a neural network to reduce the model capacity, but they ignore filters' preference for particular data when pruning the model. \cite{li2021hermes,deng2022tailorfl} allow clients with more similar data to share more identical parameters, and \cite{chen2023efficient} dynamically personalizes parameters according to batch-level information. However, they suffer from undesirable knowledge retention as depicted in Sec.\ref{sec1}.
\paragraph{Decomposition FL.}Decomposition FL can explicitly decouple knowledge by parameter decomposition \cite{hyeon2021fedpara,jeong2022factorized,mei2022resource}.\cite{jeong2022factorized} decomposes each weight into a global vector and a personalized vector. \cite{hyeon2021fedpara} proposes a novel decomposition way to address the low-rank issue. The two methods are communication-efficient but both ignore the computational costs. \cite{mei2022resource} decomposes each weight into a common part and a capacity-specific part without pruning models. However, only clients with identical capacities can perform aggregation on capacity-specific parts and personalization was not taken into consideration by them.
\section{Experiment}\label{sec5}
\subsection{Setup}\label{sec5.1}
\paragraph{Datasets.} We evaluate Pa$^3$dFL on three widely used benchmarks in FL: FashionMNIST \cite{xiao2017fashion}, CIFAR10 and CIFAR100 \cite{krizhevsky2009learning}. We partition each dataset into 100 parts respectively kept by clients. We control data heterogeneity as \cite{mcmahan2017communication,jin2023feddyn}. For CIFAR10, clients' label distributions obey $Dirichlet(\alpha=1.0)$ \cite{measure}. For CIFAR100, clients own 20 classes of data from all 100 classes. For FashionMNIST, data is i.i.d. distributed. We set the ratios of train/val/test parts as $0.8/0.1/0.1$ for each local data. 
\paragraph{Models.} We use ResNet18-GN used by \cite{jin2023feddyn} for CIFAR10, where each batch normalization layer is replaced by group normalization due to the infeasibility of batch normalization on non-I.I.D. data in FL \cite{hsieh2020non}. We use two-layer CNNs for both CIFAR100 \cite{jin2023feddyn} and FashionMNIST \cite{wang2023fedgs}. We detail the architecture of the models in Appendix \ref{apdx_c2}.
\paragraph{Capacity Setting.} We assume that each client can only afford training or testing a submodel that takes nearly $r\%$ computation costs of the full model. For width-reduction methods \cite{alam2022fedrolex, horvath2021fjord, diao2020heterofl, mei2022resource, deng2022tailorfl}, the reduction ratio $p_i$ of each client $i$ should meet $p_i^2\le r_i\%$ since the cost of a $p$-width model is $\mathcal{O}(p^2d)$ \cite{diao2020heterofl} and $d$ is the full model size. For methods that directly mask the ratio $s\%$ of parameters \cite{chen2023efficient}, $s_i\%\le r_i\%$ should hold for each client. For capacity-unaware methods, we use the maximum model that can be computed by all the users based on the minimum capacity, which enables full participation of all the local data. We conduct experiments on two settings: \textsc{Ideal} and \textsc{Hetero.}. Clients in \textsc{Ideal} settings have unlimited computation resources. In \textsc{Hetero.}, we set $r_i$ to be uniformly distributed from $1\%\sim 100\%$ across clients. For example, clients with the lowest capacity can only afford training a submodel that is nearly $\frac{1}{100}$ of the complete model.
\begin{table}
\centering
\caption{Comarison on testing accuracy (\%) of different methods w./w.o. heterogeneous capacity constraints on three datasets. The optimal result in each column is in bold font and the second optimal results are underlined.}
\label{tb_overall}
\begin{sc}
\begin{scriptsize}

\begin{tabular}{c|cc|cc|cc} 
\hline\hline
          & \multicolumn{2}{c|}{CIFAR10}      & \multicolumn{2}{c|}{CIFAR100}     & \multicolumn{2}{c}{FashionMNIST}   \\ 
\hline
 Method  & Ideal             & Hetero.             & Ideal             & Hetero.             & Ideal             & Hetero.              \\ 
\hline
FedAvg    & $75.01\pm 0.43$ & $52.42\pm 0.56$ & $38.09\pm 0.85$ & $17.11\pm 0.19$ & $\underline{91.71\pm 0.27}$ & $87.76\pm 0.48$  \\
Ditto     & $\bold{91.16\pm 0.03}$ & $83.11\pm 0.27$ & $\underline{57.07\pm 0.30}$ & $34.44\pm 0.12$ & $90.91\pm 0.35$ & $86.51\pm 0.44$  \\
pFedHN    & $83.11\pm 0.12$ & $79.33\pm 0.42$ & $32.49\pm 0.39$ & $27.38\pm 0.65$ & $84.51\pm 0.38$ & $82.51\pm 0.40$  \\ 
FLANC     & $74.64\pm 0.44$ & $53.17\pm 0.40$ & $37.84\pm 0.36$ & $23.84\pm 0.11$ & $91.43\pm 0.10$ & $85.33\pm 0.29$ \\
Fjord     & $75.94\pm 0.43$ & $71.52\pm 0.31$ & $38.51\pm 0.44$ & $34.41\pm 0.31$ & $91.04\pm 0.10$ & $\bold{90.11 \pm 0.26}$  \\
HeteroFL  & $74.17\pm 0.47$ & $68.42\pm 0.86$ & $38.97\pm 0.83$ & $26.81\pm 0.48$ & $90.56\pm 0.20$ & $85.66\pm 0.85$  \\ 
FedRolex  & $74.80\pm 0.23$ & $68.26\pm 0.31$ & $36.55\pm 0.15$ & $27.84\pm 0.25$ & $\bold{91.78\pm 0.14}$ & $87.40\pm 0.26$  \\ 
LocalOnly & $76.59\pm 0.40$ & $76.28\pm 0.17$ & $25.88\pm 0.64$ & $24.55\pm 0.05$ & $75.72\pm 0.21$ & $74.98\pm 0.07$   \\
LG-Fedavg & $81.12\pm 0.09$ & $80.39\pm 0.03$ & $33.86\pm 0.21$ & $33.75\pm 0.11$ & $78.96\pm 0.03$ & $78.21\pm 0.08$   \\
TailorFL  & $74.96\pm 0.15$ & $\underline{85.00\pm 0.84}$ & $37.98\pm 0.35$ & $41.14\pm 0.66$ & $91.63\pm 0.11$ & $88.41\pm 0.23$  \\
pFedGate  & $90.58\pm 0.10$ & $67.02\pm 3.05$ & $54.32\pm 0.62$ & $4.55\pm 0.03$  & $88.33\pm 3.90$ & $76.58\pm 1.96 $  \\ 
\hline
Pa$^3$dFL   & $\underline{91.05\pm 0.23 }$ & $\bold{86.42\pm 0.49}$ & $\bold{57.33\pm 1.24}$ & $\bold{51.48\pm 0.99}$ & $ 91.04\pm 0.42$ & $\underline{89.55\pm 0.20}$  \\
\midrule
\end{tabular}
\end{scriptsize}
\end{sc}
\end{table}

\paragraph{Baselines.} We compare Pa$^3$dFL with 3 types of methods:
\begin{itemize}
\item \textbf{Personalization.} Model sizes are limited by the lowest capacity (e.g., pFedHN \cite{shamsian2021personalized}, and Ditto \cite{li2021ditto}).
\item \textbf{Capacity-aware} Clients with different capacities train the global model without personalization (e.g.,FLANC \cite{mei2022resource}, Fjord \cite{horvath2021fjord}, HeteroFL \cite{diao2020heterofl} and FedRolex \cite{alam2022fedrolex}).
\item \textbf{Capacity-aware + Personalization.} Clients with different capacities train the global model and personalize it (e.g.,  LocalOnly \cite{mcmahan2017communication}, pFedGate \cite{chen2023efficient}, TailorFL \cite{deng2022tailorfl}, and LG-FedAvg \cite{liang2020think}.
\end{itemize} 
Details of these methods are in Appendix \ref{apdx_d3}

\paragraph{Hyperparameters.} Following \cite{jin2023feddyn}, we fix batch size $B=50$ and local epoch $E=5$ for all the datasets. We set communication rounds $R=1000/2000/500$ respectively for CIFAR10/CIFAR100/FashionMNIST. We perform early stopping when there is no gain in the optimal validation metrics for more than $0.2R$ rounds. We tune the learning rate on the grid $\{5e-3, 1e-2, 5e-2, 1e-1\}$ with the decaying rate $0.998$ and respectively tune the algorithmic hyperparameters for each method to their optimal. Each result is averaged over 3 random seeds. More details on hyperparameters are in Appendix \ref{apdx_c}.

\paragraph{Implementation.} Our experiments are conducted on a 64g Ubuntu 22.04.2 LTS server with Intel(R) Xeon(R) Silver 4314 CPU @ 2.40GHz and 4 NVidia(R) RTX3090 GPUs. Our code is realized by FL framework \cite{wang2023flgo}.
\subsection{Overall Performance}\label{sec5.2}
\begin{figure}[t]
\centering

    {%
        \includegraphics[width = .32\linewidth]{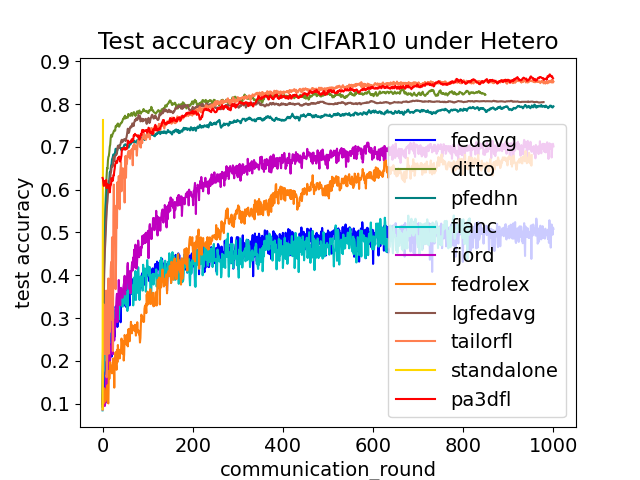}
        \includegraphics[width = .32\linewidth]{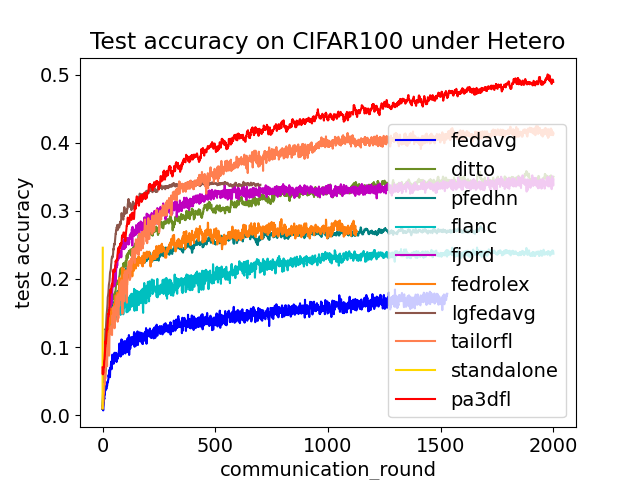}
        \includegraphics[width = .32\linewidth]{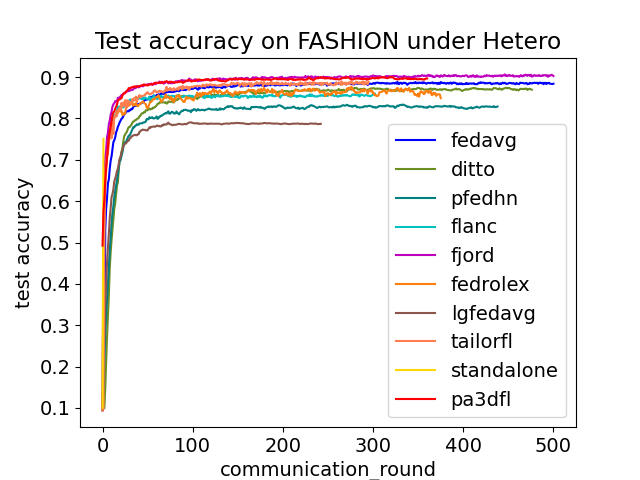}}
\caption{Test accuracy v.s. communication rounds under the \textsc{Hetero.} setting \label{fig_curve_hetero}}
\end{figure}

We first compare the overall accuracy of clients w./w.o. heterogeneous capacity constraints (e.g., \textsc{hetero.}/\textsc{ideal}) for different methods in Table \ref{tb_overall}. While existing methods suffer performance reduction from low-capacity constraints (e.g., the reduction ratios of accuracy obtained by FedAvg in CIFAR10/CIFAR100/Fashion are $30.1\%/55.0\%/4.3\%$), our proposed Pa$^3$dFL achieves the optimal results on CIFAR10 and CIFAR100 (i.e., $86.42\%$ and $51.48\%$) and competitive results against Fjord on FashionMNIST (i.e., $89.55\%$ v.s. $90.11\%$) under the same constraints. Compared with personalization-only methods (e.g., Ditto and pFedHN), Pa$^3$dFL improves model performance up to $43.3\%$ (e.g., CIFAR100-\textsc{Hetero.}) and achieves similar results in ideal cases (e.g., $91.14\%$ v.s. $91.05\%$ in CIFAR10-\textsc{ideal}). Compared with capacity-aware methods, Pa$^3$dFL consistently dominates them across all the cases in CIFAR10 and CIFAR100. We notice that our method takes no advantage in FashionMNIST-\textsc{ideal}. We attribute this result to the i.i.d. distributed data, and we still found that we outperform other personalization-only methods (e.g., Ditto and pFedHN) in this case. In addition, Pa$^3$dFL also achieves the suboptimal result in FashionMNIST-\textsc{Hetero.}, which suggests its adaptability when clients' capacity varies regardless of data distributions.
For capacity-aware methods with personalization  (e.g., LG-FedAvg, TailorFL and pFedGate), they are also able to improve model performance while maintaining different model sizes for different clients. TailorFL achieves 
similar results against Pa$^3$dFL in CIFAR10-\textsc{Hetero.} (e.g., $85.00\%$ v.s. $86.42\%$), but can only obtain a much weaker result than Pa$^3$dFL when the task becomes more difficult (e.g., $41.14\%$ v.s. $51.48\%$ in CIFAR100-UNI). Besides, TailorFL failed to personalize models for clients in all IDL cases since its personalization is based on different architectures of clients, which may limit its applicability in practice. Overall, the results in Table \ref{tb_overall} confirms the ability of Pa$^3$dFL to preserve model performance under heterogeneity in terms of data and device capacity. We further compare the learning curves of different methods in Fig.\ref{fig_curve_hetero}. Our method enjoys the fastest convergence and the highest performance across datasets.

\begin{figure}[t]
\begin{center}
\centerline{\includegraphics[width=\columnwidth]{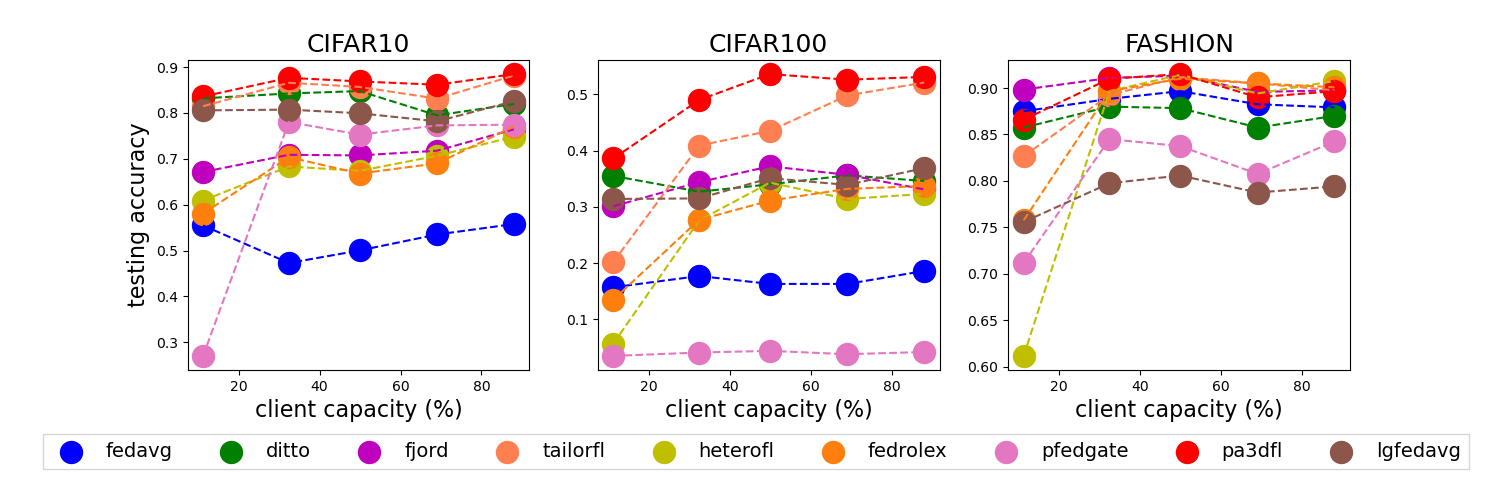}}
\caption{Model testing accuracy v.s. Client Capacity on three benchmarks}
\label{fig_cap}
\end{center}
\end{figure}
\subsection{Client Capacity-specific Performance}\label{sec5.3}
We show the capacity-specific performance under the \textsc{Hetero.} setting in Fig. \ref{fig_cap}, where we group clients' capacities into five clusters and then show the averaging testing accuracy of each cluster. The results suggest that Pa$^3$dFL outperforms other methods across different levels of client capacity in non-IID datasets (e.g. CIFAR10 and CIFAR100). We notice that TailorFL can achieve comparable results as Pa$^3$dFL in CIFAR10, however, it suffers quick performance reduction on CIFAR100 as the capacity decreases. In Fashion, Pa$^3$dFL achieves competitive results against non-personalization methods (e,g, Fjord and FedAvg) and still dominates other personalization methods. The results indicate the robustness towards different types of data distributions in practice. 
\begin{table*}
\centering
\caption{ Comparison of efficiency for different methods on CIFAR100 with the CNN model.}
\begin{footnotesize}
\label{tb_efficiency}
\begin{tabular}{c|cccc|cccc|cccc} 
\hline\hline
                  & \multicolumn{4}{c|}{FLOPs ($\times 10^7$)} & \multicolumn{4}{c|}{Peak Memory (MB)} & \multicolumn{4}{c}{Model Size ($\times 10^{-1}$MB)}  \\ 
\hline
ratio r & 1/256 & 1/64 & 1/4 & 1     & 1/256 & 1/64 & 1/4 & 1         & 1/256 & 1/64 & 1/4 & 1          \\ 
\hline
FedAvg     & - &  - &   -  &  188.0  & -  & - & -  &71.1  &    - &     - & -    & 30.4  \\
$p$-width & 3.5 & 8.2 & 59.1 & 188.0 & 23.1  &  26.2 & 45.3 &  71.1   &0.1 & 0.5 &  7.6&   30.4         \\
FLANC             & 15.0 & 18.2 & 42.3 & 87.7 & 30.1 & 34.1 & 58.4 & 90.9 &   0.6& 0.9 &7.76 & 29.3      \\
pFedGate          & 5.5 & 10.1 & 61.0 & 190.0  & 39.2 & 42.3 & 61.4 & 87.2 & 36.3 & 36.3 & 36.3 & 36.3   \\
Pa$^3$dFL           & 3.6 & 8.3 & 59.9 & 191.9 & 23.3 & 26.5 & 46.3 & 74.8  & 0.8 & 1.3 & 8.5    & 28.7      \\
\hline
\end{tabular}
\end{footnotesize}
\end{table*}
\subsection{Efficiency}\label{sec5.4}
We evaluate the efficiency of methods under the metrics FLOPs, peak memory, and model size \cite{chen2023efficient,diao2020heterofl} in Table \ref{tb_efficiency}. Results of each method except pFedGate are independently obtained by training the model on CIFAR10 with a few batches of data over 10 iterations, and the batch size is fixed to 128. For pFedGate, we evaluate its costs of the sparse model and the additional gating layer, as is detailed in Appendix \ref{apdx_pfedgate}. The term $p$-width refers to methods reducing the model along the width (e.g., Fjord, HeteroFL, FedRolex and TailorFL). Firstly, like other capacity-aware methods, Pa$^3$dFL can significantly reduce both the computation cost and communication cost of the model. Compared with $p$-width, Pa$^3$dFL brings few additional costs. However, Pa$^3$dFL saves efficiency in a level more similar to $p$-width than pFedGate and FLANC. For example, the ratio of saving amounts of FLOPs by Pa$^3$dFL and $p$-width when shrinking the model to $1/256$ are respectively $98.08\%$ and $98.13\%$ when compared to the full model used by FedAvg. We also observe similar trends of the efficiency saving by Pa$^3$dFL in the peak memory and model size. For FLANC, it can achieve much smaller FLOPs than other methods when the reduction ratio is large. However, the effectiveness of the efficiency saving is reduced as the reduction ratio becomes small. pFedGate failed to reduce the model size since full parameters might be activated during computation, resulting in the full transmission and storage of the model.
\begin{table}
\centering
\caption{The testing accuracy ($\%$) of ablation on Pa$^3$dFL.}
\label{tb_ablation}
\begin{small}
\begin{tblr}{
  cells = {c},
  vline{2} = {-}{},
  hline{1-3,5} = {-}{},
}
                       & CIFAR10 & FashionMNIST \\
Pa$^3$dFL                & $86.42$   & $89.55$        \\
w.o. HN-based Agg.      & $81.52 (-5.66\%)$   & $78.94 (-11.84\%)$      \\
FLANC decomposition     & $82.35 (-4.70\%)$   &  $84.58 (-5.54\%)$ \\ 
\end{tblr}
\end{small}
\end{table}
\subsection{Ablation Study}\label{sec5.5}
We conduct ablation studies on modules of Pa$^3$dFL to show their impacts in Table \ref{tb_ablation}. We notice that only aggregating general parameters of models without using hyper-network to aggregate personalized parts severely hurts model performance (e.g., $-5.66\%$ in CIFAR10 and $-11.84\%$ in FashionMNIST) for both i.i.d. and non-i.i.d. cases. This indicates the importance of the aggregation of personalized parts for personal knowledge fusion. When replacing the model decomposition with the one used in FLANC (i.e., Eq. (\ref{eqflanc})), our method also suffers non-neglectable performance reduction. Therefore, we claim that the parameter decomposition way is essential to the model performance since it decides how knowledge is organized in parameters as aforementioned. 

\section{Conclusion}\label{sec6}
In this work, we address the issue of model performance reduction when there exists both data heterogeneity and capacity heterogeneity in FL. We identify and tackle two key challenges to enhance the model performance when adapting the over-capacity model to clients by pruning. Different from previous works, we propose to prune the model after decomposition and develop Pa$^3$dFL to promote both general and personal knowledge sharing. We theoretically verify the effectiveness of our methods. Finally, we conduct extensive experiments to show the advantages of our proposed method in terms of accuracy and efficiency. 
We consider how to efficiently train models with rich operators and larger scalability under heterogeneous capacity constraints as our future works.

\newpage
\nocite{langley00} 
\bibliography{neurips_2024}
\bibliographystyle{unsrt}

\newpage
\appendix
\onecolumn
\section{Complexity Analysis}\label{apdx_a}
\begin{figure}[h]
\centering

    {%
        \includegraphics[width = .32\linewidth]{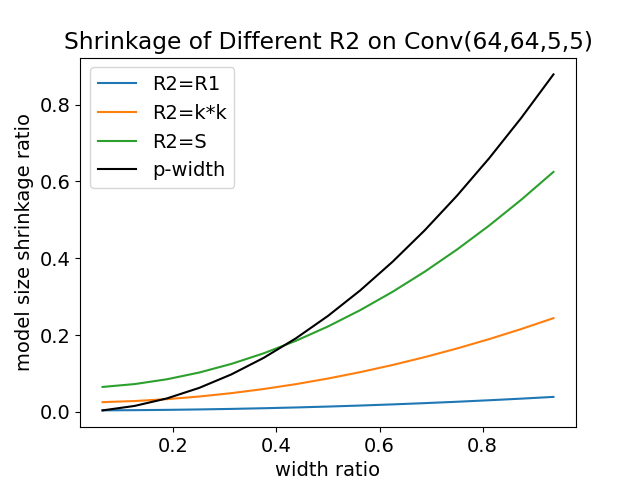}
        \includegraphics[width = .32\linewidth]{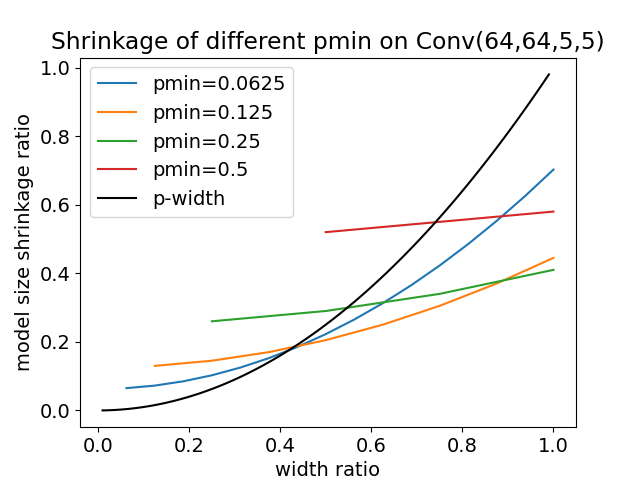}
        \includegraphics[width = .32\linewidth]{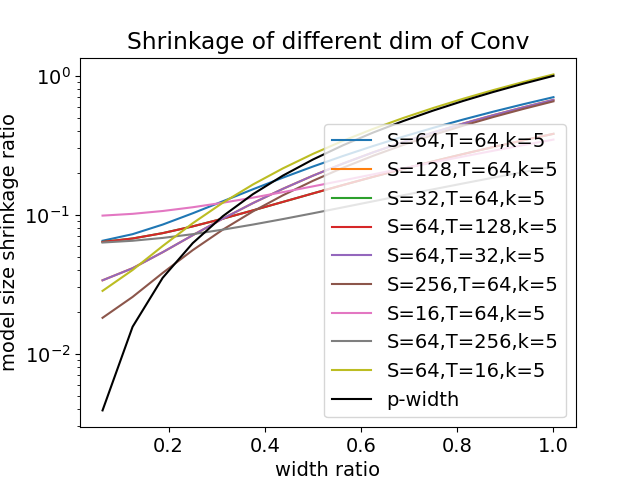}}

    {%
        \includegraphics[width = .32\linewidth]{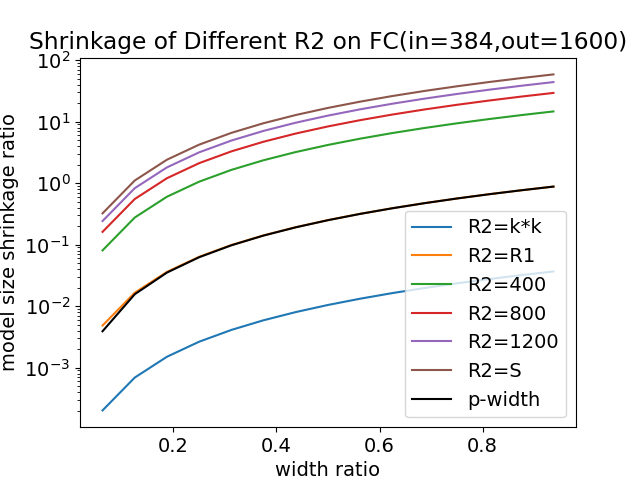}
        \includegraphics[width = .32\linewidth]{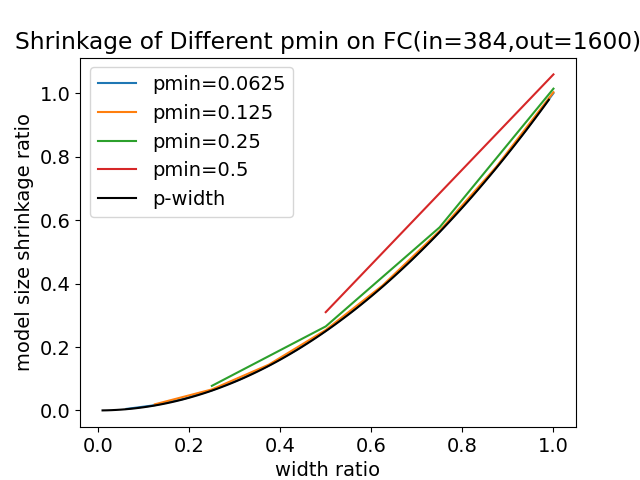}
        \includegraphics[width = .32\linewidth]{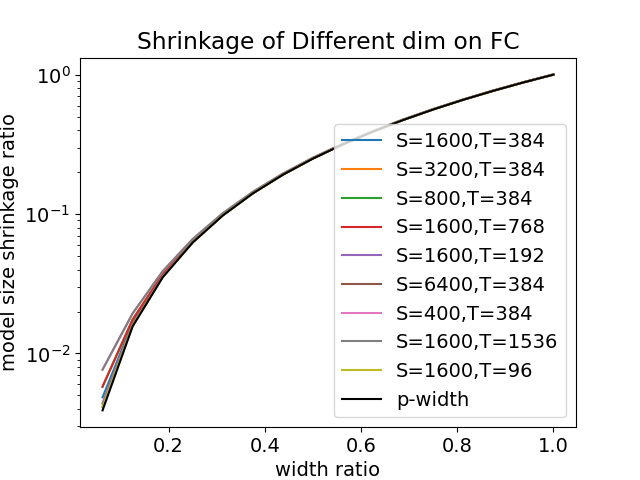}
        }
\caption{The model reduction ratios v.s. $R_2$, the lowest capacity $p_{min}$, the dimension and the type of operators\label{fig_shrinkage}}
\end{figure}

\paragraph{Model Size.} A standard convolution weight $\boldsymbol{\theta}_{conv}\in\mathbb{R}^{T\times S\times k\times k}$ is decomposed into $\boldsymbol{\theta_w}\in \mathbb{R}^{R_1k^2\times R_2}$ and $\boldsymbol{\theta_v}\in \mathbb{R}^{R_2\times \frac{ST}{R1}}$, where $T\mod R_1=0$. $k$ is the kernel size. $S$/$T$ is the number of input/output channels. $R_1$ and $R_2$ are two hyperparameters that control the amount of the convolution parameters. Therefore, the total amount of the decomposed convolution is $R_2(ST/R_1+R_1k^2)$. 
When the convolution layer is reduced by ratio $p$, the amount of corresponding parameters becomes  $R_2(\frac{p^2ST}{R1}+R_1k^2)$. 
Since $\lfloor Tp\rfloor \mod R_1=0, \forall p$ is always required in Pa$^3$dFL and a small $R_1$ may limit the expressiveness of $\boldsymbol{\theta_w}$, we always fixed $R_1=Tp_{min}$. The corresponding parameter amount is 
\begin{equation}
    R_2\left( \frac{p^2S}{p_{min}}+Tp_{min}k^2\right) = pR_2(\frac{p}{p_{min}}S + \frac{p_{min}}{p}Tk^2)
\end{equation}
We denote $p_{min}\le \sigma $ as the systemic constraint since $p_{min}$ depends on the lowest capacity of clients. The reduction ratio $t$ of a $p$-width model implemented by Pa$^3$dFL is 
\begin{equation}
    r_{Pa^3dFL} = p\left(R_2\frac{(\frac{p}{p_{min}}S + \frac{p_{min}}{p}Tk^2)}{STk^2}\right)
\end{equation}
We first visualize how the model size reduction ratio changes with the value of $R_2$ respectively on the convolution operator Conv(64,64,5,5) and Linear operator (1600, 384) that is used in CIFAR10 in Fig. \ref{fig_shrinkage}. We fixed the lowest capacity $\sigma=0.0625$ as default. Firstly, we notice that for the convolution operator, let $R_2=\max(k^2, S)$ won't significantly violate the benefit in the model reduction ratio. Even when the width ratio is small, the convolution operator brings very limited additional amounts of parameters. For the Linear operator, the model size increases quickly as $R_2$ increases, and the reduction ratio is almost consistent with $p$-width. Since $\boldsymbol{\theta_{w}}\in \mathbb{R}^{R_1\times R_2}$ where setting $R_2$ won't increase the rank of $boldsymbol{\theta_{w}}$, we set $R_2=R_1$ for all linear operators without further specification. We finally design the rule of choosing $R_1, R_2$ as follows:
\begin{equation}\label{eq_R}
    R_1=Tp_{min}, R_2=\left \{
\begin{array}{ll}
   \max(\min(S,T),k^2),& \text{if layer is convolution}\\
    R_1,& \text{if layer is linear}\\
\end{array}
\right.
\end{equation}

Under this rule, we respectively study how the reduction ratio is influenced by the systemic constraint $\sigma$ and the dimension of the operators in Fig. \ref{fig_shrinkage}. For the convolution operator, a loose constraint (i.e., a large $\sigma$) will diminish the effectiveness of model reduction when the width ratio is small. However, a small $p_{min}$ is allowed to be used under loose systemic constraints. Different dimensions of convolution layers enjoy different reduction speeds. Although additional costs are introduced by Pa$^3$dFL than $p$-width, the reduced amount of model parameters is already significant compared to the full model. For the linear operator, the results show the consistency between this rule $p$-width when the systemic constraints or the dimensions of operators vary. Further, if $\min(S, T)\ge k^2$, the reduction ratio under this rule becomes
\begin{equation}
        r_{Pa^3dFL} = p\left(R_2\frac{(\frac{p}{p_{min}}S + \frac{p_{min}}{p}Tk^2)}{STk^2}\right)\le p (\frac{1}{k^2} \frac{p}{p_{min}}+ \frac{p_{min}}{p}) = \frac{1}{k^2p_{min}}p^2+p_{min}
\end{equation}
If $p_{min}$ is small than $p^2$ and $k^2p_{min}\approx 1$, the ratio is nearly of complexity $\mathcal{O}(p^2)$. The term $\frac{1}{k^2}$ in the reduction ratio also explains why the convolution operator in Pa$^3$dFL has a larger tolerance to the value changing of $R_2$ than the linear operator in Fig. \ref{fig_shrinkage}. Therefore, one can use a hyperparameter to scale $R_2$ of convolution layers properly to enhance the rank of the common parts' matrices. Unfortunately, this rule may not work when using a super kernel size $k^2\gg \min{S,T}$, which will produce unneglectable additional costs. In our experiments, the kernel size we used won't significantly exceed the number of parameters. We remain how to specify the parameters selection rules for convolution layer with relatively large kernel sizes as our future works.
\paragraph{Training-Time Complexity.} During the model training phase, the additional complexity comes from the matrix multiplication in the parameter recovery process \cite{mei2022resource}. Given a batch of $B$ data with the feature size $q\times q$ (i.e., the height and the width) and input channel number $c_1$, a convolution operator with kernel size $k$ and output channels $c_2$ will produce $Bq^2k^2c_1c_2$ multiply-add float operations. Since the cost of weight multiplication is $k^2R_1R_2(c_2/R_1)c_1=R_2k^2c_1c_2$, the additional cost ratio is
\begin{equation}
    \frac{k^2R_1R_2(c_2/R_1)c_1=R_2k^2c_1c_2}{Bq^2k^2c_1c_2}=\frac{R_2}{Bq^2}
\end{equation}
In practice, this term is negligible when $R_2\ll Bq^2$.

\section{Convergence}\label{apdx_con}
At each communication round $t$, the server generates personal parameters from clients' embeddings by the hyper-network according to Eq. (\ref{eq_gen}) and Eq. (\ref{eq_prune})
\begin{equation}
 \{\boldsymbol{\theta}_{i, \bold v}^{(t)}\}_{i=1}^{N} = h_{gen\&prune}(\boldsymbol\theta_h^{(t)}, \bold{E}^{(t)}) = h(\boldsymbol\varphi^{(t)})
\end{equation}
where $\boldsymbol\varphi^{(t)} = (\boldsymbol\theta_h^{(t)}, \bold{E}^{(t)})$. Then, client $c_i$ will receive the model parameters $(\boldsymbol{\theta}_{\bold u}^{(t)},\boldsymbol{\theta}_{i, \bold v}^{(t)})$ and locally update them by one-step gradient descent with the step size $\eta_t$
\begin{equation}
    \boldsymbol{\theta}_{i,\bold u}^{(t+1)} = \boldsymbol{\theta}_{\bold u}^{(t)} - \eta_t\frac{\partial f_i}{\partial \boldsymbol{\theta}_{\bold u}^{(t)}},\quad \boldsymbol{\theta}_{i, \bold v}^{(t+1)} =  \boldsymbol{\theta}_{i,\bold v}^{(t)} - \eta_t\frac{\partial f_i}{\partial \boldsymbol{\theta}_{i,\bold v}^{(t)}}
\end{equation}
The server aggregates the clients' general parameters by
\begin{equation}
    \boldsymbol{\theta}_{\bold u}^{(t+1)} = \frac{1}{N}\sum_{i=1}^N \boldsymbol{\theta}_{i,\bold u}^{(t+1)} = \boldsymbol{\theta}_{\bold u}^{(t)} - \eta_t \frac{1}{N}\sum_{i}^N\frac{\partial f_i}{\partial\boldsymbol{\theta}_{\bold u}^{(t)}} = \boldsymbol{\theta}_{\bold u}^{(t)} - \eta_t \frac{\partial F}{\partial\boldsymbol{\theta}_{\bold u}^{(t)}}
\end{equation}
where $F=\frac{1}{N}\sum_{i=1}^N f_i$ denotes the global objective. The hyper-network is updated by Eq. (\ref{eq_updatehn}) 
\begin{align}
    \frac{\partial\mathcal L_{HN}}{\partial \boldsymbol\varphi^{(t)}} &= \frac{1}{N}\frac{\partial\sum_{i=1}^N \frac12 \|h(\boldsymbol\varphi^{(t)})[i] -\boldsymbol{\theta}_{i,\bold u}^{(t+1)})\|_2^2}{\partial \boldsymbol\varphi^{(t)}}\\
    &=\frac{1}{N}\sum_{i=1}^N\eta_t\frac{\partial f_i}{\partial \boldsymbol{\theta}_{i,\bold v}^{(t)}}\frac{h(\boldsymbol\varphi^{(t)})[i]}{\boldsymbol\varphi^{(t)}}\\&=\frac{\eta_t}{N}\sum_{i=1}^N\frac{\partial f_i}{\boldsymbol\varphi^{(t)}}\\&=\eta_t\frac{\partial F}{\partial \boldsymbol\varphi^{(t)}}
\end{align}
As a result, the update rule of the hyper-network is 
\begin{equation}
    \boldsymbol\varphi^{(t+1)} =  \boldsymbol\varphi^{(t)} - \gamma\eta_t\frac{\partial F}{\partial \boldsymbol\varphi^{(t)}}
\end{equation}
By viewing the general parameters and the hyper-network parameters as a whole $\boldsymbol{\theta^{(t)}} = (\boldsymbol{\theta}_{\bold u}^{(t)}, \boldsymbol{\varphi}^{(t)})$, we can directly apply the previous analysis result to our updating scheme under the following standard assumptions.
\newtheorem{assumption}{Assumption}
\begin{assumption}[Smoothness]\label{assmp:smoothness}
For each $i=1,\ldots,n$, the function $F_i$ is continuously differentiable.  
There exist constants $L_{\bold u}, L_{\boldsymbol{\varphi}}$ such that $\frac{\partial F}{\partial \boldsymbol{\varphi}^{(t)}}$ is $L_{\boldsymbol{\varphi}}$--Lipschitz with respect to~$\boldsymbol{\varphi}^{(t)}$, and for each $i=1,\ldots,n$, $\frac{\partial F}{\partial \boldsymbol{\theta}_{i, \bold u}^{(t)}}$ is $L_{\bold u}$--Lipschitz with respect to~$\boldsymbol{\theta}_{i, \bold u}^{(t)}$.
\end{assumption}

\begin{assumption}[Bounded Variance]\label{assmp:stoc-grad-var}
The stochastic gradients in Alg.\ref{alg_server} and Alg.\ref{alg_client} are unbiased and have bounded variance. That is, for all $\bold u$ and $\boldsymbol{ \varphi}$, their stochastic gradients $\boldsymbol{g}_{\bold u}^{(t)}$ and $\boldsymbol{g}_{\boldsymbol{\varphi}}^{(t)}$ have $\mathbb{E}\bigl[\boldsymbol{g}_{\bold u}^{(t)}\bigr] = \frac{\partial F}{\partial \boldsymbol{\theta}_{\bold u}^{(t)}}, \mathbb{E}\bigl[\boldsymbol{g}_{\boldsymbol{\varphi}}^{(t)}\bigr] = \frac{\partial F}{\partial \boldsymbol{\varphi}^{(t)}}$, and there exist constants $\sigma_{\bold u}$ and $\sigma_{\boldsymbol{\varphi}}$ such that 
\begin{equation}
    \mathbb{E}\bigl[ \| \boldsymbol{g}_{\bold u}^{(t)} - \frac{\partial F}{\partial \boldsymbol{\theta}_{\bold u}^{(t)}} \|_{2} \bigr] \le \sigma_{\bold u}^{2}, \mathbb{E}\bigl[ \| \boldsymbol{g}_{\boldsymbol{\varphi}}^{(t)} - \frac{\partial F}{\partial \boldsymbol{\varphi}^{(t)}} \|_{2} \bigr] \le \sigma_{\boldsymbol{\varphi}}^{2}
\end{equation}
\end{assumption}
\newtheorem{theorem_appendix}{Theorem}
\begin{theorem_appendix}[Convergence of Pa$^3$dFL]\label{theorem1}
    Suppose Assumptions~\ref{assmp:smoothness}, \ref{assmp:stoc-grad-var} hold and the step sizes $\eta$ and $\gamma$ in Pa$^3$dFL are chosen as  $\frac{2}{L_{\bold u}}$, $\frac{L_{\bold u}}{L_{\boldsymbol{\varphi}}}$ respectively, $\eta$ decays with the number of rounds $t$ and all model parameters initialized at the same point $\boldsymbol{\theta}^{(0)}$. Then, after $T$ rounds we have:
    \begin{equation}
         \frac{1}{T}(\sum_{t=0}^{T}\gamma\eta_{t}(1-\frac{\gamma\eta_{t}L_{\boldsymbol{\varphi}}}{2})\|\frac{\partial F}{\partial \boldsymbol{\varphi}^{(t)}}\|^{2} + \sum_{t=0}^{T}\eta_{t}(1-\frac{\eta_{t}L_{\boldsymbol{\varphi}}}{2})\|\frac{\partial F}{\partial \boldsymbol{\theta}_{\bold u}^{(t)}}\|^{2}) \le \frac{1}{T}(F(\boldsymbol{\theta}^{(0)}) - F^{*}) + O(\eta^{2})
    \end{equation}
    where $F^{*}$ is the lower bound of $F$.
\end{theorem_appendix}
\begin{proof}
Based on Assumptions~\ref{assmp:smoothness}, we have: 
\begin{equation}
\label{eq26}
    \begin{aligned}
        F(\boldsymbol{\theta}_{\bold u}^{(t+1)}, \boldsymbol{\varphi}^{(t+1)}) - F(\boldsymbol{\theta}_{\bold u}^{(t)}, \boldsymbol{\varphi}^{(t)}) &\le \braket{\frac{\partial F}{\partial \boldsymbol{\varphi}^{(t)}}, \boldsymbol{\varphi}^{(t+1)} - \boldsymbol{\varphi}^{(t)}} + \braket{\frac{\partial F}{\partial \boldsymbol{\theta}_{\bold u}^{(t)}}, \boldsymbol{\theta}_{\bold u}^{(t+1)} - \boldsymbol{\theta}_{\bold u}^{(t)}} 
        \\
        &+ \frac{L_{\boldsymbol{\varphi}}}{2}\|\boldsymbol{\varphi}^{(t+1)} -\boldsymbol{\varphi}^{(t)}\|^{2} + \frac{L_{\bold u}}{2}\|\boldsymbol{\theta}_{\bold u}^{(t+1)} -\boldsymbol{\theta}_{\bold u}^{(t)}\|^{2}
    \end{aligned}
\end{equation}

Consider stochastic gradient descent: $\boldsymbol{\theta}_{\bold u}^{(t+1)} = \boldsymbol{\theta}_{\bold u}^{(t)} - \eta_{t}\boldsymbol{g}_{\bold u}^{(t)}$ and $\boldsymbol{\varphi}^{(t+1)} = \boldsymbol{\varphi}^{(t)} - \gamma\eta_{t}\boldsymbol{g}_{\boldsymbol{\varphi}}^{(t)}$, based on Assumptions~\ref{assmp:stoc-grad-var} then we have:
\begin{equation}
    \begin{aligned}
    \mathbb{E}\bigl[F(\boldsymbol{\theta}_{\bold u}^{(t+1)}, \boldsymbol{\varphi}^{(t+1)}) - F(\boldsymbol{\theta}_{\bold u}^{(t)}, \boldsymbol{\varphi}^{(t)})\bigr] &\le 
    \gamma\eta_{t}(\frac{\gamma\eta_{t}L_{\boldsymbol{\varphi}}}{2} - 1)\|\frac{\partial F}{\partial \boldsymbol{\varphi}^{(t)}}\|^{2} 
    \\
    &+ \eta_{t}(\frac{\eta_{t}L_{\bold u}}{2} - 1)\|\frac{\partial F}{\partial \boldsymbol{\theta}_{\bold u}^{(t)}}\|^{2} + O(\eta^{2})
    \end{aligned}
\end{equation}
If the step sizes are selected as $\frac{2}{L_{\bold u}}$ and $\frac{L_{\bold u}}{L_{\boldsymbol{\varphi}}}$ respectively, the convergence condition is satisfied. Based on (\ref{eq26}), after $T$ rounds, we have:
\begin{equation}
    \begin{aligned}
    F^{*} - F(\boldsymbol{\theta}^{(0)}) &\le
    F(\boldsymbol{\theta}_{\bold u}^{(T)}, \boldsymbol{\varphi}^{(T)}) - F(\boldsymbol{\theta}_{\bold u}^{(0)}, \boldsymbol{\varphi}^{(0)}) 
    \\
    &\le -\gamma\sum_{t=0}^{T-1}\eta_{t}\braket{\frac{\partial F}{\partial \boldsymbol{\varphi}^{(t)}}, \boldsymbol{g}_{\boldsymbol{\varphi}}^{(t)}} -\sum_{t=0}^{T-1}\eta_{t}\braket{\frac{\partial F}{\partial \boldsymbol{\theta}_{\bold u}^{(t)}}, \boldsymbol{g}_{\bold u}^{(t)}} 
    \\
    &+ \frac{\gamma^{2}L_{\boldsymbol{\varphi}}}{2}\sum_{t=0}^{T-1}\eta_{t}^{2}\|\boldsymbol{g}_{\boldsymbol{\varphi}}^{(t)}\|^{2} + \frac{L_{\bold u}}{2}\sum_{t=0}^{T-1}\eta_{t}^{2}\|\boldsymbol{g}_{\bold u}^{(t)}\|^{2}
    \\
    \end{aligned}
\end{equation}
then:
\begin{equation}
    \begin{aligned}
        \mathbb{E}\bigl[\gamma\sum_{t=0}^{T-1}\eta_{t}\braket{\frac{\partial F}{\partial \boldsymbol{\varphi}^{(t)}}, \boldsymbol{g}_{\boldsymbol{\varphi}}^{(t)}} +\sum_{t=0}^{T-1}\eta_{t}\braket{\frac{\partial F}{\partial \boldsymbol{\theta}_{\bold u}^{(t)}}, \boldsymbol{g}_{\bold u}^{(t)}} \bigr] &= \gamma\sum_{t=0}^{T-1}\eta_{t}\|\frac{\partial F}{\partial \boldsymbol{\varphi}^{(t)}}\|^{2} +\sum_{t=0}^{T-1}\eta_{t}\|\frac{\partial F}{\partial \boldsymbol{\theta}_{\bold u}^{(t)}}\|^{2} 
        \\
        &\le F(\boldsymbol{\theta}^{(0)}) - F^{*} + \frac{\gamma^{2}L_{\boldsymbol{\varphi}}}{2}\sum_{t=0}^{T-1}\eta_{t}^{2}(\|\frac{\partial F}{\partial \boldsymbol{\varphi}^{(t)}}\|^{2} + \sigma_{\boldsymbol{\varphi}}^{2}) \\
        &+ \frac{L_{\bold u}}{2}\sum_{t=0}^{T-1}\eta_{t}^{2}(\|\frac{\partial F}{\partial \boldsymbol{\theta}_{\bold u}^{(t)}}\|^{2} + \sigma_{\bold u}^{2})
    \end{aligned}
\end{equation}
then we have:
\begin{equation}
    \begin{aligned}
        \frac{1}{T}(\sum_{t=0}^{T}\gamma\eta_{t}(1-\frac{\gamma\eta_{t}L_{\boldsymbol{\varphi}}}{2})\|\frac{\partial F}{\partial \boldsymbol{\varphi}^{(t)}}\|^{2} + \sum_{t=0}^{T}\eta_{t}(1-\frac{\eta_{t}L_{\boldsymbol{\varphi}}}{2})\|\frac{\partial F}{\partial \boldsymbol{\theta}_{\bold u}^{(t)}}\|^{2}) &\le \frac{1}{T}(F(\boldsymbol{\theta}^{(0)}) - F^{*}) + O(\eta^{2})
    \end{aligned}
\end{equation}
\end{proof}

\section{Algorithmic Details}\label{apdx_b}
\subsection{Pseudo Code of Pa$^3$dFL}\label{apdx_b1}
We respectively show the main procedure of Pa$^3$dFL on the server-side and client-side in Alg.\ref{alg_server}, Alg.\ref{alg_client}. 
\subsection{Training Tips}\label{apdx_b2}
\paragraph{Testing Model Selection.} The procedure for testing model selection is shown in Alg.\ref{alg_select}. Since we only perform the model selection for testing purposes and the process won't leak the validation data information to the training process, it won't bring risks of overfitting on the validation dataset. The model selection can be performed at any time after locally training the model for each client. Since no model training is performed, the cost of model fusion is the almost same as the inference phase. Clients can recover the model into $p$-width model first and then fuse the recovered model to further reduce the cost at the model selection phase. 
\begin{algorithm}[ht]\label{alg_select}
\caption{Pa$^3$dFL - Model Selection}
\textbf{Input}:validation data $\mathcal{D}_{val}$, models $\mathcal{M}_0$ and  $\mathcal{M}_1$, the maximum number of iterations $n$\\
\begin{algorithmic}[1]
\STATE Given the function of the model performance on the validation dataset $f: \mathcal{M}\times \mathcal{D}\rightarrow met$, set the objective function as $g(\alpha)=f(\mathcal{M}_0 + \alpha(\mathcal{M}_0-\mathcal{M}_1))$, $\alpha\in [0,1]$ 
\end{algorithmic}

\end{algorithm}
\label{alg_server}
\begin{algorithm}[ht]
\caption{Pa$^3$dFL - Server}
\textbf{Input}:clients' capacity constraints $\boldsymbol{r}$,number of clients $N$, model architecture $\mathcal{M}$, hyper-network $\mathcal{H}$, learning rate $\eta$\\
\begin{algorithmic}[1]

\PROCEDURE{Server}{$\boldsymbol{r}$}

\STATE Set minimal model width ratio $p_{min} = \min(\{p_i|i\in [N]\})$

\STATE Select $R_1,R_2$ according to Eq. (\ref{eq_R}) and specify model architecture based on $\mathcal{M}$

\STATE Initialize model common parameters $\boldsymbol{\theta_w}^{(0)}$, the hyper-network parameters $\boldsymbol{\phi}^{(0)}$, and clients' embedding $\bold U^{(0)}$

\FOR{$t\gets 0 \text{ to } T-1$} 
\STATE Randomly samples a subset $\mathcal{S}_t$ of $m$ clients
\FOR{$i\in \mathcal{S}_t$}
\STATE Generate and prune personalized parameters $\boldsymbol{\theta_v}_i^{(t)}=Prune(\mathcal{H}(\bold U^{(t)}, i), p_i)$ by its width ratio $p_i$
\STATE Send $(\boldsymbol{\theta_w}^{(t)}, \boldsymbol{\theta_v}_i^{(t)})$ to client $i$
\STATE $\boldsymbol{\theta_w}_i^{(t+1)}, \boldsymbol{\theta_v}_i^{(t+1)}=\textsc{LocalUpdate}(\boldsymbol{\theta})$
\ENDFOR
\STATE \# Aggregation
\STATE $\boldsymbol{\theta_w}^{(t+1)} = \frac{1}{m}\sum_{i\in \mathcal{S}_t}\boldsymbol{\theta_w}_i^{(t+1)}$

\STATE Compute $\mathcal{L}=\frac{1}{m}\sum_{i\in \mathcal{S}_t}\|\boldsymbol{\theta_v}_i^{(t+1)}-\boldsymbol{\theta_v}_i^{(t)}\|_2^2$ and backward
\STATE Set $\boldsymbol{\phi}^{(t+1)} =\boldsymbol{\phi}^{(t)}-\eta \frac{\partial \mathcal{L}}{\partial\boldsymbol{\phi}^{(t)}}$
\STATE Set $\boldsymbol{U}^{(t+1)} =\boldsymbol{U}^{(t)}-\eta \frac{\partial \mathcal{L}}{\partial\boldsymbol{U}^{(t)}}$
\ENDFOR
\ENDPROCEDURE

\end{algorithmic}
\end{algorithm}
\label{alg_client}
\begin{algorithm}[tb]


\caption{Pa$^3$dFL - Client}
\label{alg:1}
\textbf{Input}: hyperparameter $\lambda$, model architecture $\mathcal{M}$, local epochs $E$, batch size $B$, learning rate $\eta$\\
\begin{algorithmic}[1]
\PROCEDURE{LocalUpdate}{$\boldsymbol{\theta}$}
\STATE Set $\boldsymbol{\theta}_{\bold w, head}, \boldsymbol{\theta}_{\bold w, enc}, \boldsymbol{\theta}_{\bold v, head}, \boldsymbol{\theta}_{\bold v, enc}= \boldsymbol{\theta}$ 
\STATE Freeze parameters of $\boldsymbol{\theta}_{\bold w, head}$
\FOR{$e\gets 1 \text{ to } E$}
\FOR{batch data $ (x,y)$ in local data}
\STATE Recover $p$-width model from parameters $\boldsymbol{\theta}_{enc} = \mathcal{R}(\boldsymbol{\theta}_{\bold w, enc}, \boldsymbol{\theta}_{\bold v, enc})$
\STATE Compute representation $x'=\mathcal{M}_{p, enc}(x;\boldsymbol{\theta}_{enc})$
\STATE Compute loss on the fixed global head $\mathcal{L}_{g} = \ell(\boldsymbol{\theta}_{\bold w, head}(x'), y)$
\STATE Compute loss on the local head with detached representations $\mathcal{L}_{c} = \ell(\boldsymbol{\theta}_{\bold w, head}(\text{detach}(x')), y)$
\STATE Compute regularization term $\mathcal{L}_{reg}$ according to Eq. (\ref{eq_reg})
\STATE Compute loss $\mathcal{L}=\mathcal{L}_{g}+\mathcal{L}_c+\lambda\mathcal{L}_{reg}$
\STATE Backward and update model parameters
\ENDFOR
\ENDFOR
\\
\RETURN $\boldsymbol{\theta}_{\bold w}, \boldsymbol{\theta}_{\bold v}$
\ENDPROCEDURE
\end{algorithmic}
\end{algorithm}
\subsection{Regularization Term}
We follow \cite{mei2022resource} to add a similar orthogonal regularization term to enhance the expressiveness of the decomposed parameters as follows:
\begin{equation}\label{eq_reg}
    \mathcal{L}_{reg} = \lambda\sum_{i\in \{\text{conv layer}\}}\|\boldsymbol{S_i} - diag(\boldsymbol{S_i})\|^2, \boldsymbol{S_i}=\boldsymbol{\theta_{w,i}}^\top \boldsymbol{\theta_{w,i}}
\end{equation}

where $diag(\cdot)$ only preserves the diagonal elements of the input and $\lambda$ is the hyper-parameter. This regularization is only made for convolution operators.
Although we have a different scheme with FLANC \cite{mei2022resource} in the parameter recovery process, the orthogonal regularization term they used is also suitable to our method. In FLANC, columns of the common parts $\boldsymbol{\theta_{w}}$ are viewed as the neural basis of the parameters of each filter. Therefore, they regularize the basis to be orthogonal to each other to increase the expressiveness by 
\begin{equation}
    \mathcal{L}_{FLANC,reg} = \frac{1}{L}\sum_{l=1}^{L}\|\boldsymbol{\theta_{w}}^\top\boldsymbol{\theta_{w}} - \boldsymbol{I}\|_2^2
\end{equation}
where $L$ denotes the number of decomposition layers in the model. In Pa$^3$dFL, columns in each basic component construct the basis of filters in the final weights that are delivered from the component. Therefore, encouraging columns to be orthogonal to each other also helps increase the expressiveness of filters in our parameter recovery scheme. Since the corresponding kernel size for linear operators is $1$, requiring the orthogonality of scalars among columns of a basic component is meaningless. Therefore, we only perform this regularization term for convolution operators instead of all the operators like FLANC. In addition, we notice that the term used by FLANC will force the $l_2$ norm of each column in $\boldsymbol{\theta_{w}}$ to be closest to one. We eliminate this issue by ignoring the diagonal elements in the product in Eq. (\ref{eq_reg}) to further enhance model performance from empirical observations.
\section{Experimental Details}\label{apdx_c}

\subsection{Datasets}\label{apdx_c1}

\begin{figure}
\centering
\subfigure[CIFAR10]{
	\label{fig_cifar10}
	\includegraphics[width = .3\linewidth]{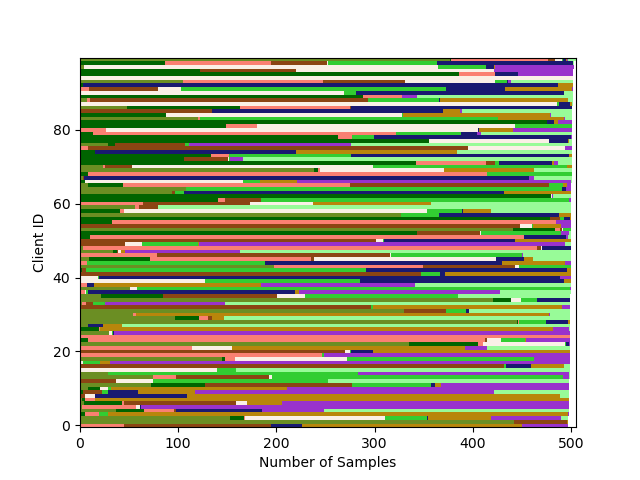}}
\subfigure[CIFAR100]{
	\label{fig_cifar100}
	\includegraphics[width = .3\linewidth]{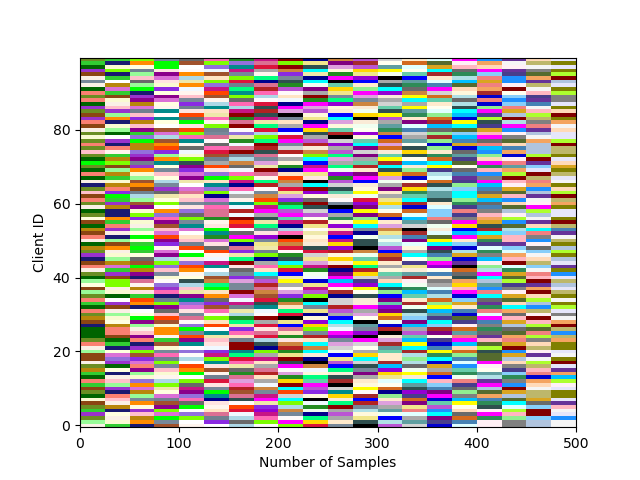}}
\subfigure[FashionMNIST]{
	\label{fig_fashion}
	\includegraphics[width = .3\linewidth]{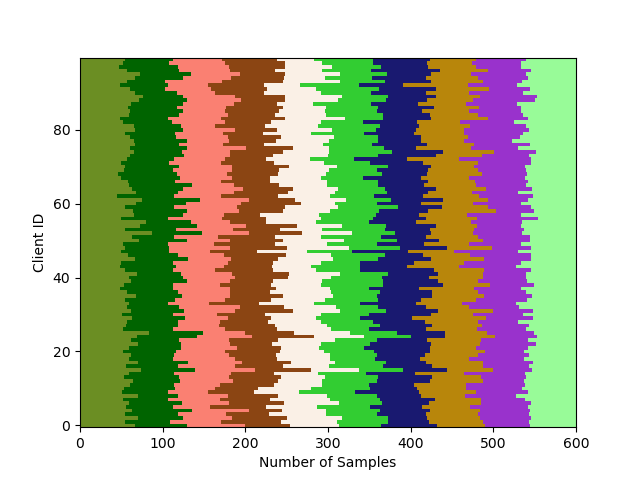}}
\caption{The visualization of data partition for CIFAR10 (a), CIFAR100 (B) and FashionMNIST (C). Each bar in the figures represents a client's local dataset and each label is assign to one color. The length of each bar reflects the size of the local data.}
\label{fig:7}
\end{figure}

\paragraph{CIFAR10\textbackslash100} Each CIFAR dataset \cite{krizhevsky2009learning} consists of total 60000 32x32 colour images in 10\textbackslash100 classes (i.e., 50000 training images and 10000 test images). We use random data augmentation on the two datasets \cite{mcmahan2017communication}. For CIFAR10, We partition the training data by letting the local label distribution $\bold p_k\sim Dirichlet(\alpha \bold p*)$ for each client, where $\bold p*$is the label distribution in the original dataset. We use $\alpha=1.0$ in our experiments.  For CIFAR100, we randomly let each client to own 20 classes of 100 classes. We provide the visualized partition in Fig. \ref{fig_cifar10} and \ref{fig_cifar100}.

\paragraph{FashionMNIST.} The dataset \cite{xiao2017fashion} consists of a training set of 60,000 examples and a test set of 10,000 examples, where each example is a $28\times28$ size image of fashion and associated with a label from 10 classes. In this work, we partition this dataset into 100 clients in a i.i.d. manner and equally allocate the same number of examples to each one. A direct visualization of the partitioned result is provided in Fig. \ref{fig_fashion}.
\subsection{Models}\label{apdx_c2}
For all the three datasets, we use the classical CNN architecture used by \cite{mcmahan2017communication}. The information on the models is concluded in Table \ref{tb_model}.

\begin{table}[]
    \centering
    \begin{tabular}{c|ccc}
    \hline
         & CIFAR10&CIFAR100&FACHION \\
    \hline
         Input Shape&(3,32,32)&(3,32,32)&(1,28,28)\\
    \hline
         \multirow{4}*{Conv Layer}&(3, 64, kernel=7, stride=2, pad=3)&(3,64,kernel=5), &(1,32,kernel=5, pad=2), \\
         &pool, relu&pool, relu&pool, relu
         \\
         &ResBlock(2) *4&(64,64,kernel=5),&(32,64,kernel=5, pad=2), \\
         &&pool, relu&pool, relu
         \\
    \hline
         \multirow{3}*{FC Layer}&(512, 10)&(1600,384), relu &(3136,512), relu \\
         &&(384,192), relu &(512,128), relu  \\
         &&(192,100), relu &(128,10), relu \\
    \hline
    \end{tabular}
    \caption{The model architecture.}
    \label{tb_model}
\end{table}

\subsection{Baselines \& Hyperparameters}\label{apdx_d3}
We consider the following baselines including SOTA personalized and capacity-aware methods in this work
\begin{itemize}
    \item \textsc{LocalOnly} is a non-federated method where each client independently trains its local model;
    \item \textsc{FedAvg} \cite{mcmahan2017communication} is the classical FL method that iteratively averages the locally trained models to update the global model;
    \item \textsc{Ditto} \cite{li2021ditto} personalizes the local model by limiting its distance to the global model for each client with a proximal term. We tune the coefficient $\mu$ from $\{0.01, 0.1, 1\}$ for this method;
    \item \textsc{pFedHN} \cite{shamsian2021personalized} is a SOTA PFL method that uses a hyper-network to generate personalized parameters for each client;
    \item \textsc{HeteroFL} \cite{diao2020heterofl} is a capacity-aware FL method that prunes model parameters in a nested manner and uses the Scaler to align information across models of different capacities.
    \item \textsc{Fjord}\cite{horvath2021fjord} drops filters of a neural network at each layer in order and it aligns the information across models with different scalability by knowledge distillation.
    \item \textsc{FLANC}\cite{mei2022resource} reduces the model size by decomposing the model parameters into commonly shared parts and capacity-specific parts. They average the commonly shared parts to update the model, and only capacity-specific parts at the same capacity level can be aggregated. In addition, it uses an orthogonal regularization term to enhance the expressiveness of commonly shared parts. We tune the term's coefficient $\lambda$ from $\{0.01, 0.1, 1.0\}$
    \item \textsc{TailorFL}\cite{deng2022tailorfl} is an efficient personalized FL method that prunes filters in a personalized manner, which enhances the local model performance by encouraging similar clients to share more identical filters. It regularizes the local training objectives by limiting the norm of the importance layer proposed by them. We tune the coefficient of the regularization from $\{0.1, 0.5\}$.
    \item \textsc{pFedGate}\cite{chen2023efficient} dynamically masks blocks of model parameters according to batch-level information to personalize the model parameters. A local gating layer is maintained by each client locally to decide which parameters should be activated during computation. We set the default splitting number $B=5$ as they did and we tune this method by varying the learning rate of the gating layer from the grid $\{0.005, 0.01, 0.05, 0.1\}$ of the model learning rate.
    \item 
    \textsc{LG-FEDAVG}\cite{liang2020think} is a method that jointly learns compact local representations on each device and a global model across all devices.

    \item \textsc{FEDROLEX}\cite{NEURIPS2022_bf5311df} employs a rolling sub-model extraction scheme that allows different parts of the global server model to be evenly trained, which mitigates the client drift induced by the inconsistency between individual client models and server model architectures.
\end{itemize}

\paragraph{Hyperparameters.}
 we fix batch size $B=50$ and local epoch $E=5$ for all the datasets. We set communication rounds $R=1000/2000/500$ respectively for CIFAR10/CIFAR100/FashionMNIST. We perform early stopping when there is no gain in the optimal validation metrics for more than $0.2R$ rounds. We tune the learning rate on the grid $\{0.005, 0.01, 0.05, 0.1\}$ with the decaying rate $0.998$ and respectively tune the algorithmic hyperparameters for each method to their optimal. For Pa$^3$dFL, we select the dimensions of clients' embeddings and hidden layers of HN from $\{(64, 64), (16, 128)\}$, and we tune the depth of HN encoder over $\{2,3,4\}$ , the regularization coefficient from $\{0.1, 0.01, 0.001\}$ and the learning rate of HN from $\{1.0, 0.1, 0.01, 0.001\}$. We conclude the optimal configurations for all the methods in Table \ref{tb_con_cifar10}.
 \begin{table}
 \label{tb_con_cifar10}
\centering
\caption{Optimal Configuration on CIFAR10}
\begin{tblr}{
  cells = {c},
  cell{1}{2} = {c=2}{},
  cell{1}{4} = {c=2}{},
  hline{1-3,15} = {-}{},
}
          & \textbf{Ideal}     &                      & \textbf{Hetero.}   &                      \\
          & \textbf{Step size} & \textbf{Algorithmic} & \textbf{Step Size} & \textbf{Algorithmic} \\
FedAvg    & 0.1                & -                    & 0.1                & -                    \\
Ditto     & 0.1                & 0.1                  & 0.05               & 1.0                  \\
pFedHN    & 0.1                & {[}5, 100, 2]        & 0.1                & {[}5, 100, 2]        \\
FLANC     & 0.1                & 0.01                 & 0.1                & 0.01                 \\
Fjord     & 0.1                & -                    & 0.1                & -                    \\
HeteroFL  & 0.1                & -                    & 0.1                & -                    \\
FedRolex  & 0.1                & -                    & 0.1                & -                    \\
LocalOnly & 0.005              & -                    & 0.005              & -                    \\
LG-FedAvg & 0.005              & -                    & 0.1                & -                    \\
TailorFL  & 0.1                & 0.1, 0.05            & 0.1                & 0.5, 0.05            \\
pFedGate  & 0.1                & 0.005                & 0.1                & 0.1                  \\
Pa$^3$dFL    & 0.1                & 0.1,0.001,[64,64,4]  & 0.1                & 0.1,0.001,[64,64,3]   
\end{tblr}
\end{table}

\begin{table}\label{tb_con_cifar100}
\centering
\caption{Optimal Configuration on CIFAR100}
\begin{tblr}{
  cells = {c},
  cell{1}{2} = {c=2}{},
  cell{1}{4} = {c=2}{},
  hline{1-3,15} = {-}{},
}
          & \textbf{Ideal}     &                      & \textbf{Hetero.}   &                      \\
          & \textbf{Step size} & \textbf{Algorithmic} & \textbf{Step Size} & \textbf{Algorithmic} \\
FedAvg    & 0.1                & -                    & 0.1                & -                    \\
Ditto     & 0.1                & 0.1                  & 0.05               & 1.0                  \\
pFedHN    & 0.05               & {[}5, 100, 2]        & 0.1                & {[}5, 100, 2]        \\
FLANC     & 0.1                & 1.0                  & 0.1                & 0.01                 \\
Fjord     & 0.1                & -                    & 0.1                & -                    \\
HeteroFL  & 0.05               & -                    & 0.05               & -                    \\
FedRolex  & 0.1                & -                    & 0.1                & -                    \\
LocalOnly & 0.1                & -                    & 0.1                & -                    \\
LG-FedAvg & 0.1                & -                    & 0.1                & -                    \\
TailorFL  & 0.1                & 0.5, 0.2             & 0.01               & 0.5, 0.2             \\
pFedGate  & 0.1                & 0.05                 & 0.1                & 0.05                 \\
Pa$^3$dFL    & 0.1                & 0.001,1.0,[16,128,2] & 0.1                & 0.01,0.1,[16,128,3]  
\end{tblr}
\end{table}
\begin{table}\label{tb_con_fashion}
\centering
\caption{Optimal Configuration on FASHION}
\begin{tblr}{
  cells = {c},
  cell{1}{2} = {c=2}{},
  cell{1}{4} = {c=2}{},
  hline{1-3,15} = {-}{},
}
          & \textbf{Ideal}     &                      & \textbf{Hetero.}   &                      \\
          & \textbf{Step size} & \textbf{Algorithmic} & \textbf{Step Size} & \textbf{Algorithmic} \\
FedAvg    & 0.1                & -                    & 0.05               & -                    \\
Ditto     & 0.05               & 1.0                  & 0.05               & 1.0                  \\
pFedHN    & 0.1                & -                    & 0.1                & -                    \\
FLANC     & 0.1                & 1.0                  & 0.1                & 1.0                  \\
Fjord     & 0.1                & -                    & 0.1                & -                    \\
HeteroFL  & 0.05               & -                    & 0.1                & -                    \\
FedRolex  & 0.1                & -                    & 0.1                & -                    \\
LocalOnly & 0.1                & -                    & 0.1                & -                    \\
LG-FedAvg & 0.1                & -                    & 0.1                & -                    \\
TailorFL  & 0.1                & 0.1, 0.2             & 0.01               & 0.05, 0.2            \\
pFedGate  & 0.1                & 0.1                  & 0.1                & 0.05                 \\
Pa3dFL    & 0.1                & 0.01,0.1,[64,64,3]   & 0.1                & 0.001,1.0,[64,64,4]  
\end{tblr}
\end{table}

\subsection{Efficiency Evaluation For pFedGate}\label{apdx_pfedgate}
We discuss the details of the efficiency evaluation for pFedGate. While the efficiency of pruning-based baselines can be directly computed by tracing the memory and the intermediates during computation, this manner fails to evaluate the efficiency of pFedGate since dynamically masking model parameters (i.e., setting their values to zero) without actually dropping them out from the memory won't save any computation efficiency. \cite{chen2023efficient} simply assumes the computation cost is proportion to the size of the non-masked model parameters and implements pFedGate based on zero-masking. As illustrated above, this type of implementation cannot be directly evaluated by tracing the systemic variables (e.g memory). A possible way is to split each weight in a model into independently stored parts, which allows users to drop specified parts before feeding data into the model. However, it's too complex to realize such a mechanism in practice due to the varying splitting numbers, operator shapes,  model architectures, and dimension alignment. As a result, we keep the consistency of our implementation with the authors' official open-source codes\footnote{https://github.com/yxdyc/pFedGate}. Since the additional costs of pFedGate usually come from the gating layer, we respectively consider the cost of the gating layer and the sparse model, and we then regard the sum of them to obtain the evaluated results. For a sparse model with sparsity $s$, it shares almost the same computation cost as the pruning-based method with width reduction ratio $\sqrt{s}$. The gating layer receives a batch of data and outputs a vector with the legnth of the total number of blocks, which is irrelevant to the reduction ratio of the model and is only decided by the input shapes and the model splitting manner. Therefore, we evaluate the additional cost of the gating layer as the difference between the cost of a full ordinary model and the cost of the pFedGate's mdoel with sparsity 1.0 for each batch size. Then, we use the estimated costs of the gating layer as the additional costs for all reduction ratios when batch size is specified. We finally add the cost of the shrunk model and the cost of the gating layer together to obtain the total cost of pFedGate.
\section{Additional Experiments}\label{apdx_d}
\begin{table}
\centering
\caption{ Comparison of efficiency for different methods on CIFAR100 with the CNN model.}
\label{tb_full_efficiency}
\begin{center}

\begin{tblr}{
  cell{1}{1} = {c=2}{},
  cell{1}{3} = {c=4}{},
  cell{1}{7} = {c=4}{},
  cell{2}{1} = {c=2}{},
  cell{3}{1} = {r=4}{},
  cell{7}{1} = {r=4}{},
  cell{11}{1} = {r=4}{},
  cell{15}{1} = {r=4}{},
  vlines,
  hline{1-3,7,11,15,19} = {-}{}
}
                &       & \textbf{FLOPs} ($\times 10^7$) &        &        &       & \textbf{Peak Memory} (MB) &       &       &       \\
reduction ratio &       & 1/256  & 1/64   & 1/4    & 1     & 1/256       & 1/64  & 1/4   & 1     \\
p-width         & B=16  & 0.4    & 1.0    & 7.3    & 23.5  & 17.0        & 17.4  & 20.4  & 25.7  \\
                & B=64  & 1.7    & 4.1    & 29.5   & 94.0  & 19.2        & 20.8  & 31.4  & 45.1  \\
                & B=128 & 3.5    & 8.2    & 59.0   & 188.1 & 23.1        & 26.2  & 45.3  & 71.1  \\
                & B=512 & 14.2   & 32.8   & 236.3  & 752.3 & 41.2        & 53.4  & 127.9 & 227.3 \\
FLANC           & B=16  & 1.8    & 2.2    & 5.2    & 10.9  & 17.9        & 18.4  & 22.0  & 28.2  \\
                & B=64  & 7.4    & 9.0    & 21.1   & 43.8  & 22.8        & 24.8  & 38.3  & 55.2  \\
                & B=128 & 14.9   & 18.1   & 42.2   & 87.7  & 30.1        & 34.1  & 58.4  & 90.9  \\
                & B=512 & 59.9   & 72.7   & 169.1  & 350.9 & 69.6        & 84.0  & 179.6 & 305.4 \\
pFedGate        & B=16  & 0.6    & 1.2    & 7.6    & 23.7  & 27.6        & 28.1  & 31.0  & 36.4  \\
                & B=64  & 2.7    & 5.0    & 30.5   & 95.0  & 32.4        & 33.9  & 44.6  & 58.2  \\
                & B=128 & 5.5    & 10.1   & 61.0   & 190.0 & 40.8       & 43.9 & 63.0 & 88.8  \\
                & B=512 & 22.1   & 40.7   & 244.1  & 760.1 & 68.0        & 80.2  & 154.7 & 254.1 \\
Pa$^3$dFL         & B=16  & 0.47   & 1.0    & 8.0    & 27.3  & 17.1        & 17.6  & 21.3  & 28.6  \\
                & B=64  & 1.8    & 4.2    & 30.2   & 97.8  & 19.4        & 21.0  & 32.4  & 48.0  \\
                & B=128 & 3.6    & 8.3    & 59.9   & 191.9 & 23.3        & 26.5  & 46.3  & 74.8  \\
                & B=512 & 14.5   & 33.3   & 237.9  & 756.1 & 41.1        & 53.9  & 128.9 & 231.0 
\end{tblr}
\end{center}
\end{table}
\subsection{Full Efficiency Comparison}\label{apdx_d1}
We provide a full comparison of the computation efficiency of different methods by varying batch sizes in Table \ref{tb_full_efficiency}. When using a small batch size and a small reduction ratio, the reduced amount on both FLOPs and peak memory of Pa$^3$dFL is closest to that of p-width. We notice that if batch size is small and the reduction is large, pFedGate achieves a larger reduced amount in computation costs. However, since clients with large reduction ratios usually have more computation costs,  it will be more useful to save more efficiency for clients with small reduction ratios rather than large ones.
\begin{figure}[t]
\centering

    {%
        \includegraphics[width = .32\linewidth]{fig_curve_cifar10_uni.png}
        \includegraphics[width = .32\linewidth]{fig_curve_cifar100_uni.png}
        \includegraphics[width = .32\linewidth]{fig_curve_fashion_uni.png}}

    {%
        \includegraphics[width = .32\linewidth]{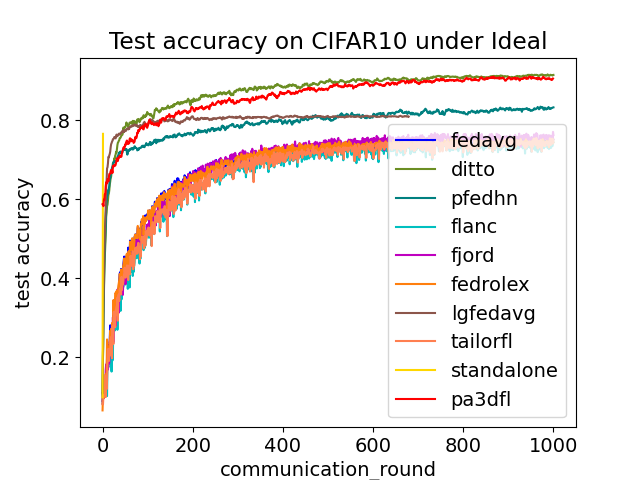}
        \includegraphics[width = .32\linewidth]{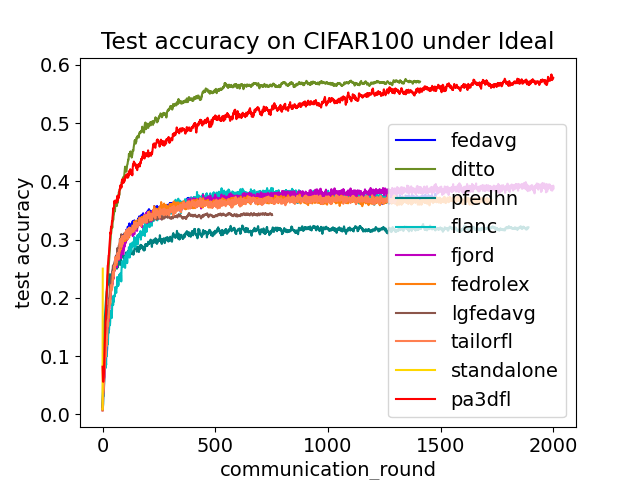}
        \includegraphics[width = .32\linewidth]{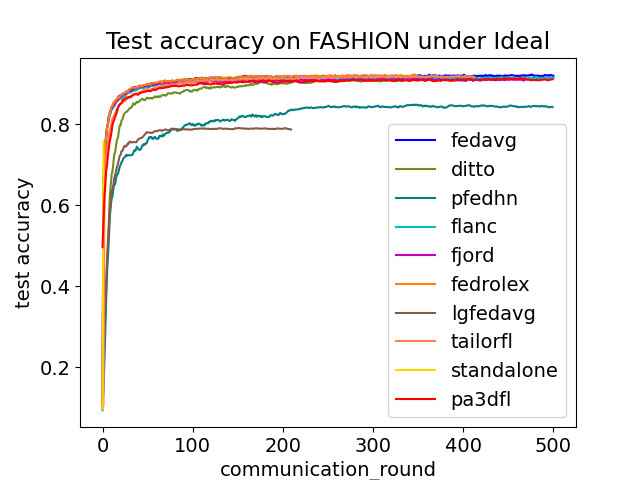}
        }
\caption{Test accuracy v.s. communication rounds under the setting UNI (the above row) and IDL (the bottom row)\label{fig_curve}}
\end{figure}
\subsection{Learning Curve}\label{apdx_d2}
\label{apdx_learning_curve}
We visualize the learning curve of different methods in Fig. \ref{fig_curve}. In \textsc{Hetero.} setting, Pa$^3$dFL consistently outperforms other methods in CIFAR10 and CIFAR100 and achieves competitive results against the SOTA method Fjord. In \textsc{Ideal} setting, Pa$^3$dFL still achieves outstanding performance regardless of data distributions. We also notice that our method does not converge until 2000 rounds in CIFAR100 while already achieving the highest performance among all the methods, indicating the room for accelerating convergence of our method to further enhance its performance. We will investigate this issue in our future work.
\subsection{Impact of Regularization}\label{apdx_exp_reg}
\begin{figure}[t]
\centering

    {%
        \includegraphics[width = .32\linewidth]{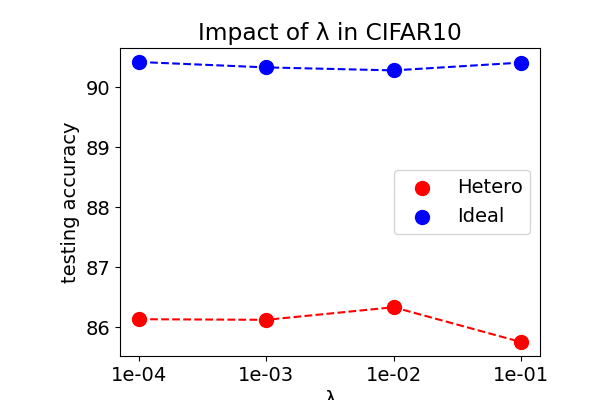}
        \includegraphics[width = .32\linewidth]{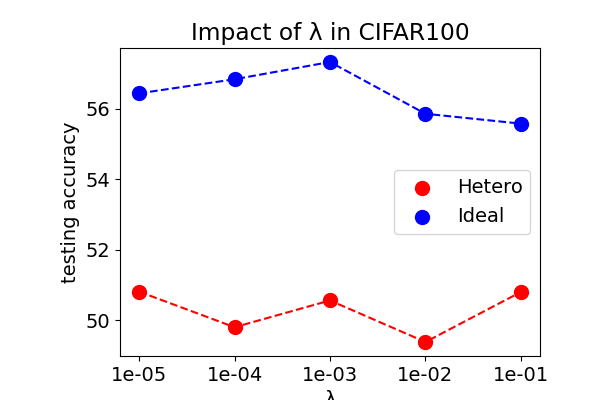}
        \includegraphics[width = .32\linewidth]{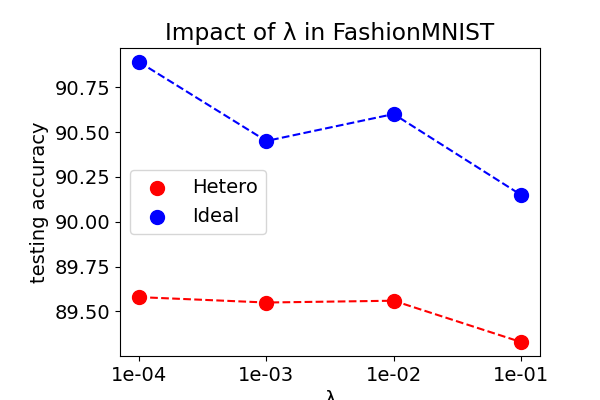}}

\caption{Test accuracy v.s. Regularization coefficient\label{fig_lmbd}}
\end{figure}
\begin{figure}[t]\label{fig_hndep}
\centering

    {%
        \includegraphics[width = .35\linewidth]{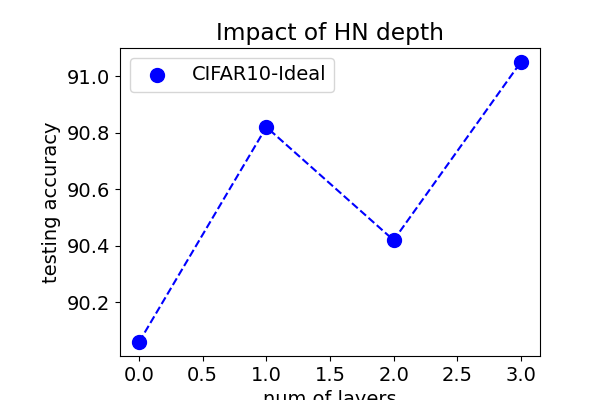}
        \includegraphics[width = .35\linewidth]{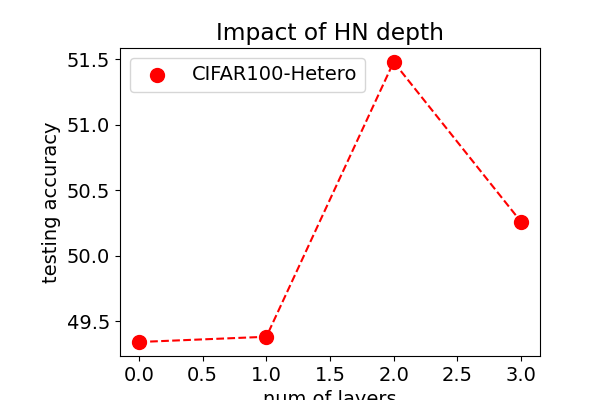}}

\caption{Test accuracy v.s. HN depth\label{fig_lmbd}}
\end{figure}
We conduct the impact of the regularization term in Fig. \ref{fig_lmbd} by varying the value of the coefficient $\lambda$ on three datasets for both the ideal case and the hetero. case. In FashionMNIST, a small regularization coefficient is preferred. We infer that this is due to the knowledge has been already well organized in model parameters. The impact of different values of the regularization coefficient is not obvious in CIFAR10 and needs to be carefully tuned in practice. These results also suggest $\lambda=1e-3$ can be a proper choice without prior knowledge.
\begin{table}[]\label{tb_hncost}
\centering
\caption{ Generating/optimization time cost (s) of hyper-networks with different numbers of layers for 20 clients.}
\begin{tabular}{llll}
\hline
                 & 1-layer   & 2-layer   & 3-layer   \\ \hline
CNN-CIFAR100  & 0.01/0.26 & 0.01/0.27 & 0.01/0.30 \\ \hline
ResNet18-CIFAR10    & 0.03/0.81 & 0.03/0.85 & 0.03/0.87 \\ \hline
\end{tabular}
\end{table}
\section{Computation Cost of HN}
We evaluate the computation cost of generating personal parameters by HN and optimization of HN on CIFAR10 (e.g., ResNet18) and CIFAR100 (2-layer CNN) in Table \ref{tb_hncost} with a (64,64,layers) HN. The time cost of hyper-network optimization and parameter generation are both extremely small when compared to local training. In addition, scaling the model also brings limited additional cost, indicating the adaptability of HN in practice.

\section{Impact of HN depth}
Generally, the clients' model performance can benefit from a deep HN as shown in Fig.\ref{fig_hndep}. 

\section{Limitation}
There exist several limitations of our work. First, We only evaluate Pa$^3$dFL on two commonly used operators, e.g., convolution operator and linear operator.  A more broad range of operators like attention are not discussed in this work. However, Pa$^3$dFL can also be extended other operators since we compress the model from a fine-grained matrix level. We will investigate the extension of Pa$^3$dFL to other operators (e.g., attention, rnn) in our future work. Second, we have only validated the effectiveness of our proposed method on CV models and datasets. How Pa$^3$dFL will perform on other types of tasks is unexplored.  In addition, Pa$^3$dFL's performance and convergence speed should be further improved when there exists no capacity constraints. Finally, we manually search the architecture of proper hyper-networks for each task, which is tedious and introduces large computation costs in network architecture searching. Nevertheless, we confirm the effectiveness of generating personalized parameters using hyper-networks. We will improve these mentioned issues in our future work.

\end{document}